\documentclass[letterpaper, 10 pt, conference]{ieeeconf}  

\IEEEoverridecommandlockouts                              

\usepackage{pgfplotstable}
\usepackage{multirow}
\usepackage{graphicx}
\usepackage{subfig}
\usepackage[font=footnotesize,labelfont=bf]{caption}
\usepackage{float}
\usepackage{mathtools}
\usepackage{color}
\usepackage{amssymb}
\usepackage{cuted}
\usepackage{verbatim}
\usepackage{tabularx}
\usepackage{colortbl}
\definecolor{mylightgreen}{RGB}{155,247,166}

\title{LISNeRF Mapping: LiDAR-based Implicit Mapping via Semantic Neural Fields for Large-Scale 3D Scenes}

\author{Jianyuan Zhang$^{1}$, Zhiliu Yang$^{1}$ and Meng Zhang$^{2}$ \thanks{$^{1}$ J. Zhang and Z. Yang are with the School of Information Science and Engineering, Yunnan University, Kunming, Yunnan, China}
\thanks{$^{2}$ M. Zhang is with the TUM School of Computation, Information and Technology (CIT), Technical University of Munich, 80333 Munich, Germany}
}

\begin{document}

 
\maketitle
\begin{strip}
\vspace{-4.5em}
\begin{center}

\centering
\setlength{\tabcolsep}{0.2em}
\setlength{\fboxsep}{0pt}
{\renewcommand{\arraystretch}{1.5}
\vspace{-0.7cm}
\begin{tabular}{c c c c }
\includegraphics[width=0.24\linewidth]{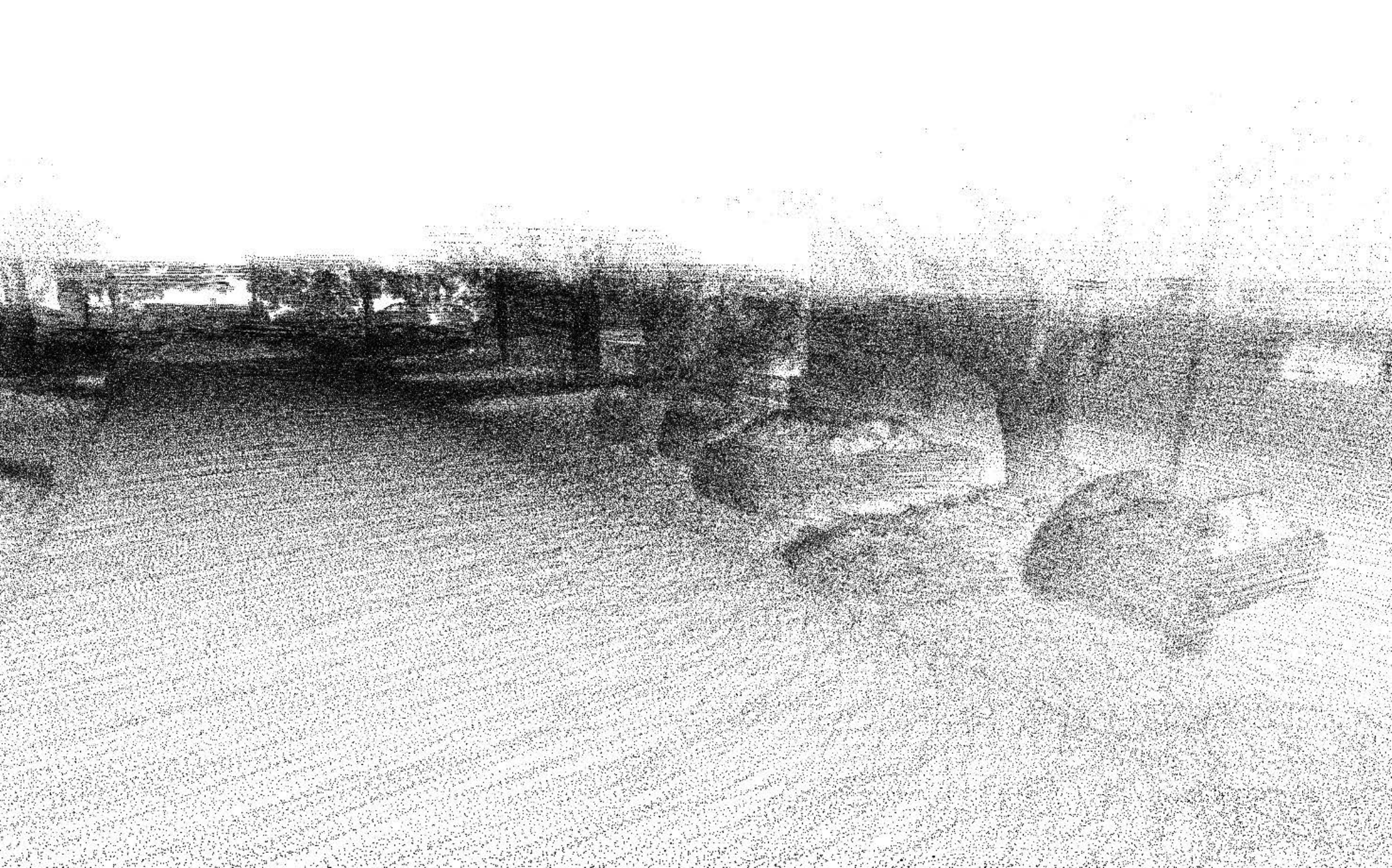} &
\includegraphics[width=0.24\linewidth]{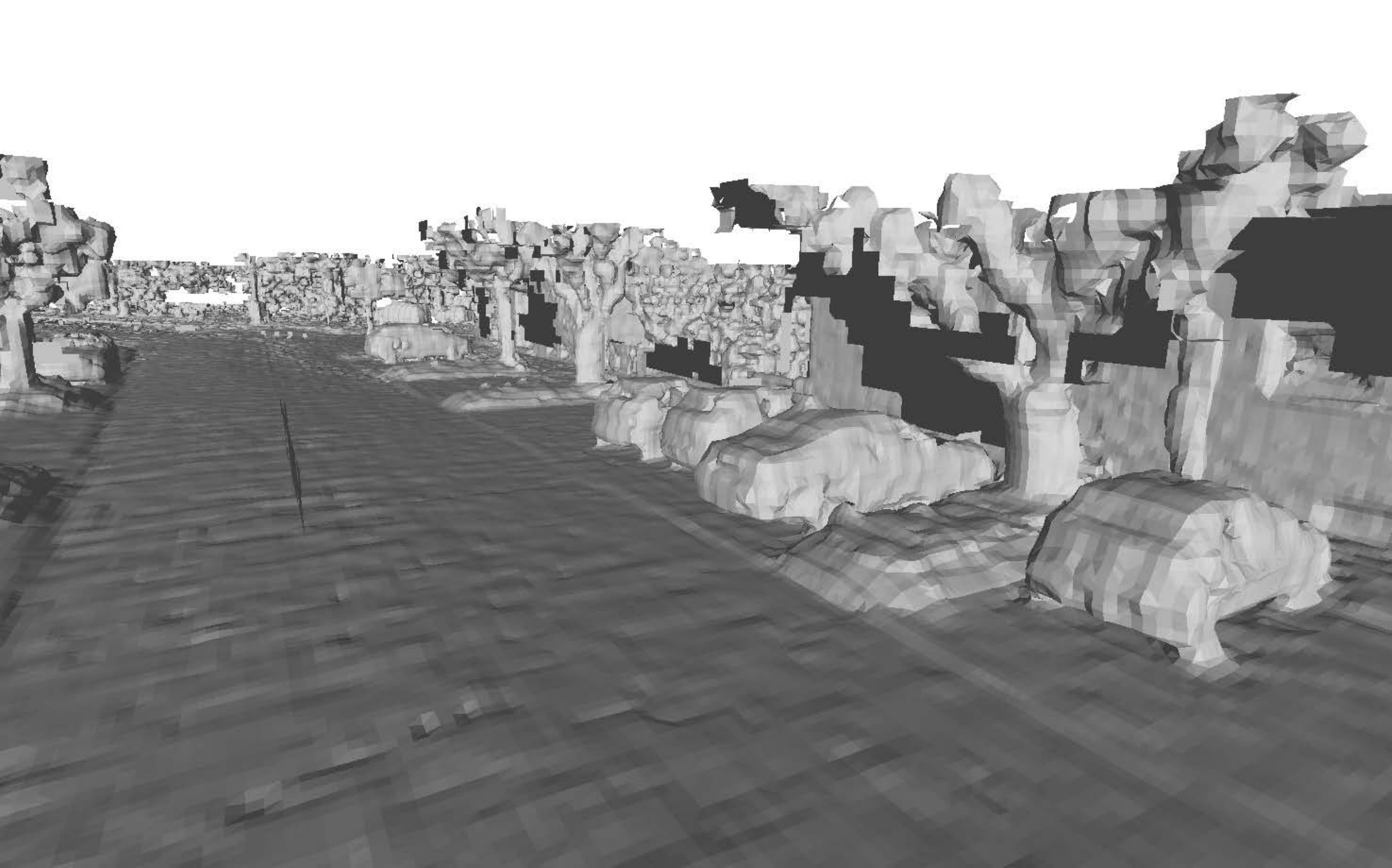} & 
\includegraphics[width=0.24\linewidth]{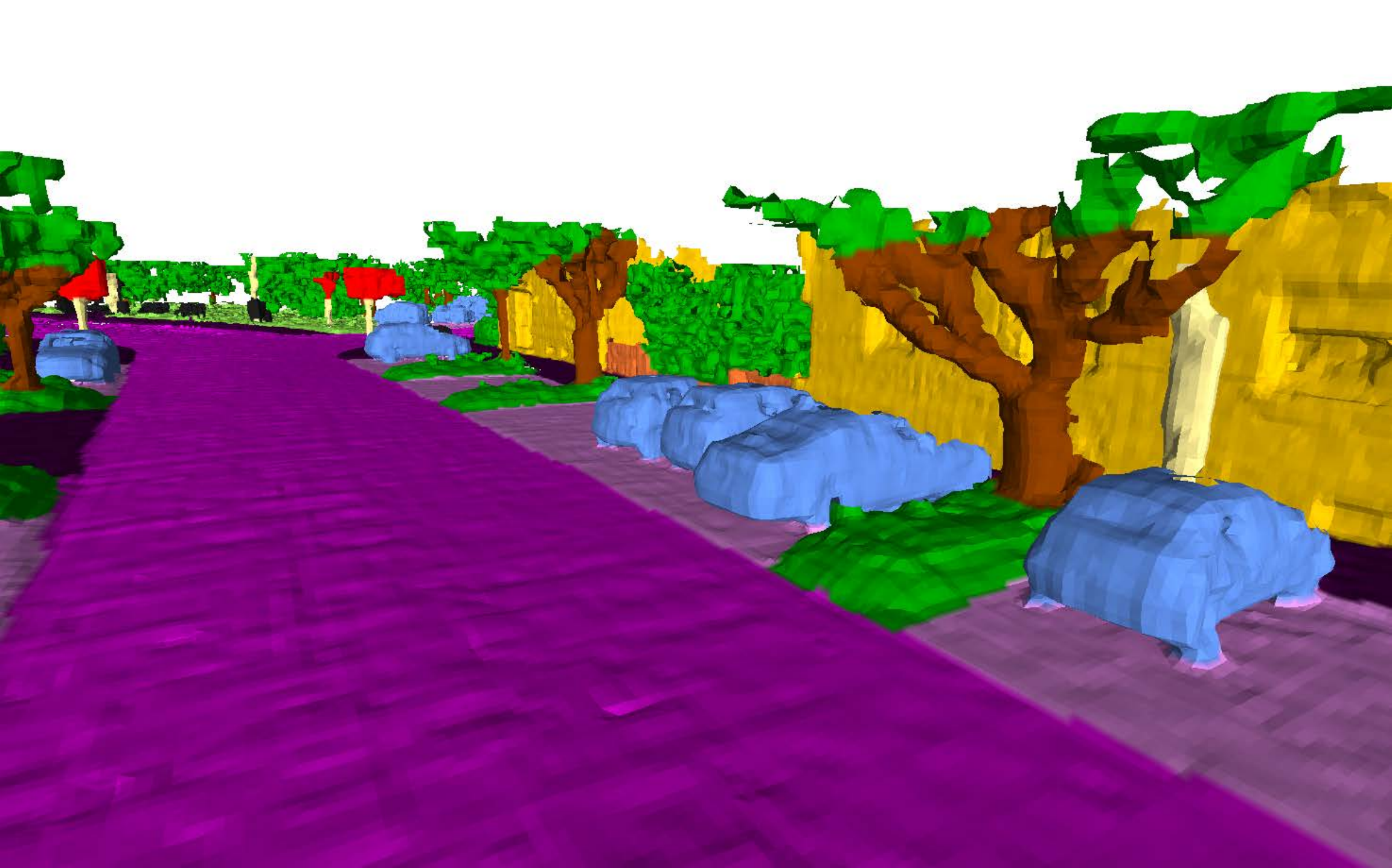}&
\includegraphics[width=0.24\linewidth]{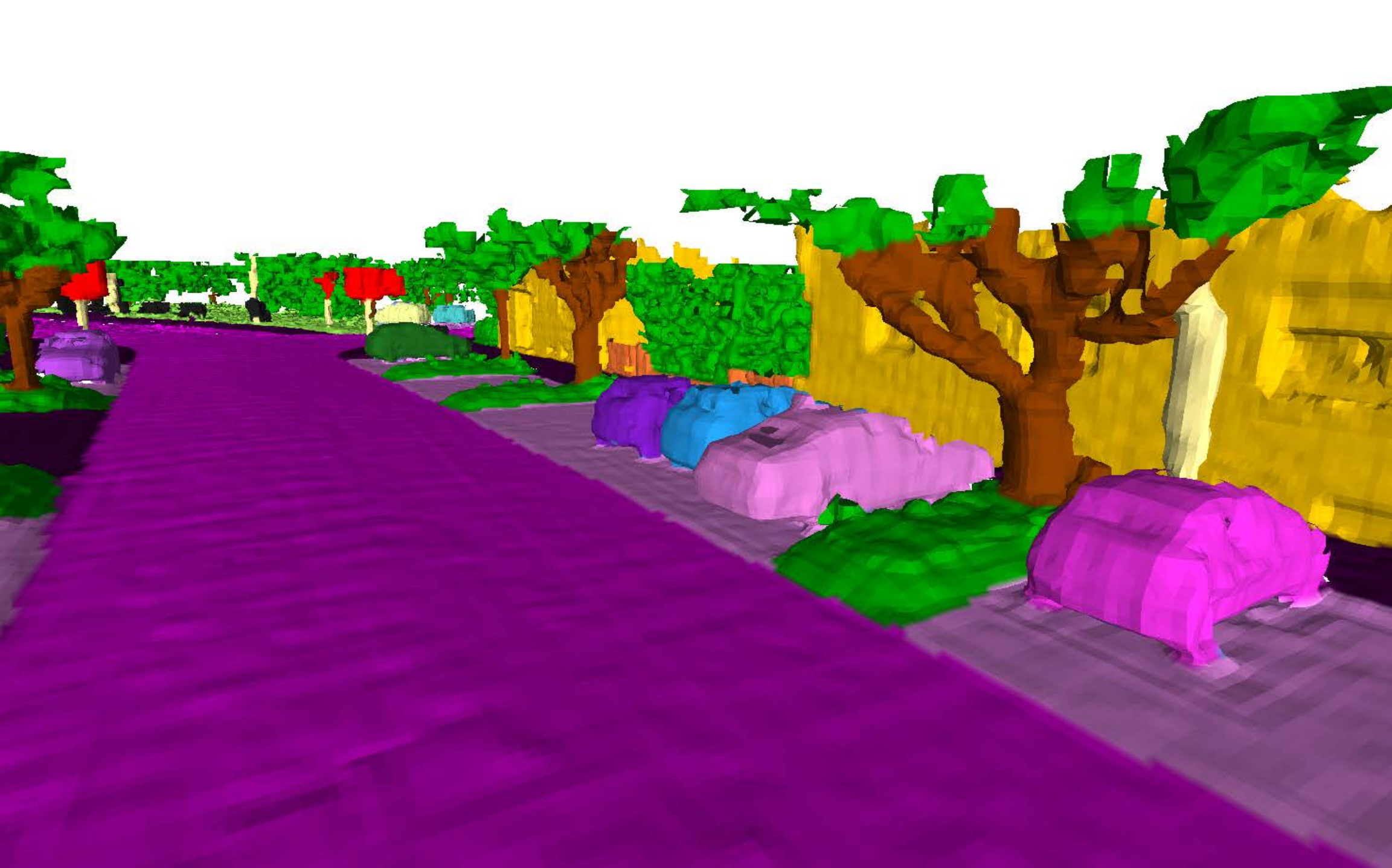}\\
(a) Point Cloud Input& (b) Neural Geometric Mesh & (c) Ours (Neural Semantic)& (d) Ours (Neural Panoptic)
\end{tabular}}
\footnotesize
\captionof{figure}{Our LISNeRF framework implicitly represents a semantic scene with LiDAR-only input. Different mapping results are shown above via evaluating on sequence 00 of SemanticKITTI. (a) is the raw LiDAR point cloud for input. (b) is the neural scene representation using an implicit method called SHINE\_Mapping \cite{zhong2023shine}. (c) and (d) are the semantic and panoptic reconstruction results via our implicit mapping.}\label{fig:datasetEx}

\end{center}%
\vspace{-0.5cm}
\end{strip}

\thispagestyle{empty}
\pagestyle{empty}



\begin{abstract}
Large-scale semantic mapping is crucial for outdoor autonomous agents to fulfill high-level tasks such as planning and navigation. This paper proposes a novel method for large-scale 3D semantic reconstruction through implicit representations from posed LiDAR measurements alone. We first leverage an octree-based and hierarchical structure to store implicit features, then these implicit features are decoded to semantic information and signed distance value through shallow Multilayer Perceptrons (MLPs). We adopt off-the-shelf algorithms to predict the semantic labels and instance IDs of point clouds. We then jointly optimize the feature embeddings and MLPs parameters with a self-supervision paradigm for point cloud geometry and a pseudo-supervision paradigm for semantic and panoptic labels. Subsequently, categories and geometric structures for novel points are regressed, and marching cubes are exploited to subdivide and visualize the scenes in the inferring stage. For scenarios with memory constraints, a map stitching strategy is also developed to merge sub-maps into a complete map. Experiments on two real-world datasets, SemanticKITTI and SemanticPOSS, demonstrate the superior segmentation efficiency and mapping effectiveness of our framework compared to current state-of-the-art 3D LiDAR mapping methods. 
\end{abstract}

\section{INTRODUCTION}
Mapping and localization in unknown environments is the key technology for autonomous driving and robots' roaming. On the one hand, in certain complex environments like urban street areas, map is a prerequisite for path planning, and semantic-enriched maps enable the feasibility of intelligent navigation. On the other hand, accurate 3D semantic reconstruction of scene facilitates the virtual simulation of auto-vehicles and mobile robots.

Most existing works of 3D semantic reconstruction \cite{mascaro2022volumetric} \cite{cartillier2021semantic} \cite{rosinol2020kimera} are developed for RGB-D cameras and indoor environments, which are not suitable for large-scale outdoor environments. For large-scale mapping, LiDAR plays a crucial role due to its ability to provide accurate distance measurement. However, large-scale 3D dense reconstruction and semantic mapping based on LiDAR sensor remains challenging because of the sparsity of LiDAR point clouds and the huge size of outdoor environments.



Recently, Neural Radiance Fields (NeRF) \cite{mildenhall2021nerf} has shown promising results in implicitly reconstructing a scene via adopting a Multi-Layer Perceptron \cite{ortiz2022isdf}, \cite{zhu2022nice}, \cite{sucar2021imap}. These NeRF-based frameworks are able to achieve higher mapping quality in indoor environments with a RGB-D sensor. While, these indoor-oriented methods face challenges when they are extended to large-scale environments, due to sensor detection range and computation resources. SHINE\_Mapping \cite{zhong2023shine} creatively employs sparse grids to implicitly represent a large-scale urban scene based on the LiDAR data alone, but it only provides geometric modeling of the environment and lacks of semantic elements, which is indispensable for modern high-level autonomous tasks.  


To this end, we propose a novel implicit LiDAR mapping framework to incorporate semantic elements into the dense map (See Fig. \ref{fig:datasetEx}). Inspired by \cite{zhong2023shine}, we exploit sparse octree-based feature embeddings to implicitly represent and store the semantic information. These feature embeddings are obtained by optimizing loss functions with self-supervision from point cloud, and pseudo-supervision from semantic segmentation or panoptic segmentation. Given the spatial location coordinates of an arbitrary point, the Signed Distance Function (SDF) value is inferred through Geometric Neural Fields (GNF) and the semantic label is inferred through Semantic Neural Field (SNF).
For explicitly displaying the implicit scene, Marching Cubes algorithm is adopted to reconstruct a scene with the form of semantic mesh. Besides, semantic mapping paradigm is further extended to the panoptic mapping by stacking more MLPs.

In summary, our contributions are as follows:
\begin{itemize}
    \item We propose a novel method to construct neural implicit semantic map for large-scale environment from posed LiDAR data alone.
    \item We also extend neural semantic mapping to neural panoptic mapping by leveraging off-the-shelf panoptic segmentation modules. Then triple MLPs are stacked into the mapping for regressing instance labels.
    \item We demonstrate a map concatenation strategy and a dynamic objects elimination method, showing the feasibility of deploying LISNeRF in large-scale and real-world mapping tasks.
\end{itemize}

 \begin{figure*}[t!]
    \centering

    \includegraphics[width=2\columnwidth]{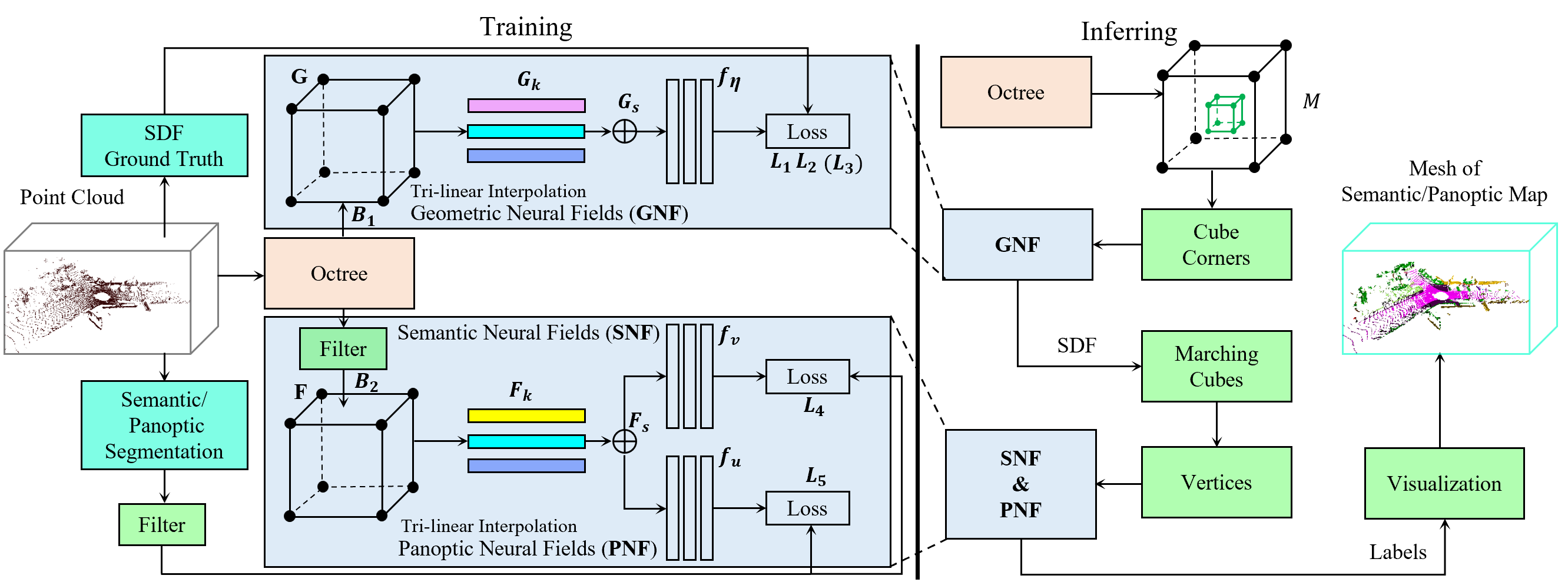}
    \caption{\textbf{The overview of our LISNeRF framework}. We first construct the Geometry Neural Fields (GNF), Semantic Neural Fields (SNF) and Panoptic Neural Fields (PNF) with the learnable feature embeddings of octree and MLPs in the training procedure. To be noticed, we filter out sampled points falling in free space and only use surface points for the training of SNF and PNF. We then calculate the current map size $M$ and predict the semantic labels of cube vertices in the inferring procedure. Given any arbitrary cube corner point in the map, we query the SDF value through GNF. After obtaining the vertices using the Marching Cubes algorithm, we exploit SNF and PNF to query the their semantic labels. Specifically, semantic mapping obtains the semantic label using the entire feature embedding through SNF. Panoptic mapping utilizes feature embeddings with a certain proportion of length to regress the classes of `thing and stuff' through PNF, and leverages other part of feature embeddings to obtain instance ID of the `thing' classes through PNF.}
    \vspace{-0.2cm}
    \label{fig:sysoverview}
\end{figure*}

\section{RELATED WORK}
\textbf{Explicit mapping.} Traditional 3D reconstruction represents a scene explicitly, and lots of map presentations are developed, such as point cloud (LeGO-LOAM \cite{shan2018lego}, Loam\_livox \cite{lin2020loam}), surfels (SuMa \cite{behley2018efficient}), occupancy grids (Cartographer \cite{hess2016real}), triangle meshes ( Puma \cite{vizzo2021poisson}, BnV-fusion \cite{li2022bnv}). Kimera \cite{rosinol2020kimera} provides a semantic mapping method that only requires running on CPU. SuMa++ \cite{chen2019suma++} extends SuMa \cite{behley2018efficient} to semantic mapping, which is able to eliminate dynamic objects in the environment. However, surfel is a type of discrete representation, the map built with surfels is relatively sparser compared to triangle mesh. Voxblox \cite{oleynikova2017voxblox} utilizes Truncated Signed Distance Function (TSDF) to reconstruct a dense map in real-time. Voxblox++ \cite{grinvald2019volumetric} incorporates semantic information into Voxblox framework, but it only fits RGB-D sensor so that is infeasible for large-scale mapping.

\textbf{Implicit mapping with color images or RGB-D data.} Contrast to the explicit scene representation, recent works leveraging neural implicit scene representation, such as NGLOD \cite{takikawa2021neural}, Di-fusion \cite{huang2021di}, and NeuralRecon \cite{sun2021neuralrecon}, achieve significant success. These methods allow for synthesizing novel perspectives and generating photo-realistic rendering results. As summed up in TABLE \ref{tab:relatedwork}, NICE-SLAM \cite{zhu2022nice} can incrementally reconstruct a scene, but it is limited to the indoor environment. Semantic-NeRF \cite{zhi2021place} introduces an extra semantic head to represent a semantic scene for indoor environments. PNF \cite{kundu2022panoptic} builds a panoptic radiance field that supports panoptic segmentation, view synthesis and scene editing. 

\textbf{Implicit mapping with LiDAR data.} Large-scale urban-level 3D reconstruction highly depends on LiDAR to provide precise range detection. As shown in TABLE \ref{tab:relatedwork}. SHINE\_Mapping \cite{zhong2023shine} employs an octree-based method to implicitly represent a large-scale scene, which is memory-efficient. NeRF-LOAM \cite{deng2023nerf} realizes large-scale 3D outdoor reconstruction and localization with remarkable precision. Efficient LNeRF \cite{yan2023efficient} implicitly represents a large-scale scene with hash code and only requires a few minutes for training. 
However, all these algorithms are short of semantic information, impeding their applications for high-level tasks.

\textbf{Semantic segmentation of point cloud.} Semantic segmentation of point cloud data is mainly divided to two approaches. (i). Apply 3D convolutions to the point cloud for semantic segmentation, such as Pointnet++ \cite{qi2017pointnet++}, Cylinder3D \cite{zhou2020cylinder3d}, DGPolarNet \cite{song2022dgpolarnet}. (ii). Project 3D point cloud into 2D range images and utilize traditional CNNs for semantic segmentation, such as SqueezeSegV2 \cite{wu2019squeezesegv2}, RangeNet++ \cite{milioto2019rangenet++}. Besides, panoptic segmentation frameworks, such as MaskPLS \cite{marcuzzi2023mask}, Panoptic-PHNet \cite{li2022panoptic}, 4D-PLS \cite{aygun20214d}, outputs the classes for stuff and things temporally consistent instance identities (IDs) are generated for things. Different from previous NeRF-based LiDAR mapping methods which only focused on refining the geometry, our method is able to integrate off-the-shelf point clouds segmentation techniques to realize semantic and panoptic mapping.


\begin{table}[h]
\centering
\caption{\textbf{Summary of the most related mapping works}. "Li" stands for LiDAR sensors, "C" stands for cameras or RGB-D sensors, "I" stands for implicit representation, "E" stands for explicit representation. ("Rep"= Representation of the scene,  "Sem" = Semantic segmentation, "Pan" = Panoptic segmentation, "LargeS" = Large-scale mapping.)}
\begin{tabular}{ c| c c c c c }
\hline
Methods & Rep & Sensors & Sem & Pan & LargeS \\
\hline
Kimera \cite{rosinol2020kimera}         & E & C &   \checkmark         &            & \checkmark \\
Puma \cite{vizzo2021poisson}            & E & Li &            &            & \checkmark \\
Suma++ \cite{chen2019suma++}            & E & Li & \checkmark &            & \checkmark \\
NICE-SLAM \cite{zhu2022nice}            & I & C &            &            & \\
iSDF \cite{ortiz2022isdf}               & I & C &            &            & \\
iMAP \cite{sucar2021imap}               & I & C &            &            & \\
Semantic-NeRF \cite{zhi2021place}       & I & C & \checkmark &            & \\
PNF \cite{kundu2022panoptic}            & I & C & \checkmark & \checkmark & \checkmark \\
NeRF-LOAM \cite{deng2023nerf}           & I & Li &            &            & \checkmark \\
Efficient LNeRF \cite{yan2023efficient} & I & Li &            &            & \checkmark\\
SHINE\_Mapping \cite{zhong2023shine}    & I & Li &            &            & \checkmark \\
\hline
Ours                                    & I & Li & \checkmark & \checkmark & \checkmark\\
\hline
\end{tabular}
\label{tab:relatedwork}
\end{table}

\section{METHODOLOGY}
Our goal is to design a novel framework to implicitly represent a semantic scene built from LiDAR-only input for large-scale environments. Overview of our framework is given in Fig. \ref{fig:sysoverview}.


\subsection{Implicit Semantic Mapping}
We implicitly represent a scene with both of geometric and semantic information. Geometry is represented via SDF value, and semantic information is assigned with objects category label. Instance ID is also assigned for panoptic mapping scenario, as is illustrated in Fig. \ref{fig:sysoverview}. All SDF values and category labels are stored in the grids of octree.

\subsubsection{\textbf{Octree-based Grids}}
Similar to NGLOD \cite{takikawa2021neural}, we  first store feature embeddings in a sparse voxel octree (SVO). One octree grid is made up by 8 corners. Each corner contains 2 one-dimensional feature embeddings ($G$ and $F$) with different lengths. $G$ stores SDF value, and $F$ stores semantic label and instance ID. The details of $G$ and $F$ are further depicted in section \ref{subsec:GEOFeature} and section \ref{subsec:SematFeature}. Different from NGLOD \cite{takikawa2021neural} which stores every level of octree features, our method only store the last $L$ levels of octree features. The level of octree is named as $Level \; k$, where $k$ = 0, 1, 2, ... , $L$-1. To be noticed, the last level is defined as $Level \; 0$. The purpose of this pruning operation is to optimize memory usage for the large-scale scene construction. Besides, hash table is applied for fast grid querying, two hash tables $B_1 = \{ \{G_0\}, \{G_1\}, ... ,\{G_{L-1}\} \}$ 
and $B_2 = \{ \{F_0\}, \{F_1\}, ... ,\{F_{L-1}\} \}$ are created to  store geometry features and semantic features respectively. Set $\{G_k\}$ and set $\{F_k\}$ stand for the entire sets of feature embeddings within level $k$. When $G_k$ and $F_k$ for certain corners are needed, the index of corners for corresponding octree grids are retrieved inside $B_1$ and $B_2$.

Same to SHINE\_Mapping \cite{zhong2023shine}, Morton code is implemented to convert 3D poses to 1D vector for fast retrieving the key of hash table. To locate the boarders of map, coordinates of maximum grid $max_k \in \mathbb{R}^3$ and minimum grid $min_k \in \mathbb{R}^3$ are recorded for each level $k$ of octree. The granularity of map is selected to fit the memory limitation of mapping tasks, and the size of current map can be calculated in each level $k$, which is:
\begin{equation}
M = (max_k - min_k)/s_{cube}
\end{equation}
where $s_{cube}$ denotes the size of Marching Cubes, $M \in \mathbb{R}^3$ means the cubes number of the whole map in the $x$, $y$ and $z$ axis.

\subsubsection{\textbf{Geometry Features Construction}}
\label{subsec:GEOFeature}
The signed distance value (SDF) \cite{takikawa2021neural} is the signed distance between a point $x_i$ to its closest surface. 
In order to obtain the SDF ground truth for a sampled point, iSDF \cite{ortiz2022isdf} examines its distance to every beam endpoints in the same point clouds batch, and choose the minimum distance as the SDF value for supervision in the training. For efficiency, we directly calculate the distance between sampled point  $x_i$ and endpoint along the same beam as supervision signal, skipping the searching procedures inside the batch.

As introduced above, ${G}$ is entire set of 1D feature embedding, and the length of single feature embedding, $G \in \mathbb{R}^{H_1}$, is $H_1$. $G$ is randomly initialized with Gaussian distribution and will be optimized in the training stage. The procedures to implicitly construct SDF value are depicted as follow:
\begin{itemize}    
\item Given an arbitrary point in the space, Morton coding is applied to convert the point's 3D coordinates to 1D code, which is used to locate the grids of corresponding level of octree. 

\item To certain level of octree, we retrieve eight $G$ of eight corners of corresponding grids through hash table $B_1$ and obtain the $G_k$ of the point via executing tri-linear interpolation.

\item We query the $G_k$ up to last three levels of octree. From coarse-grained level to fine-grained level, whose order is $Level \; 2$, $Level \; 1$, $Level \; 0$. Then, three feature embeddings, $G_2$, $G_1$ and $G_0$, are generated.

\item The concatenated feature embedding, $G_S = G_2 + G_1 +G_0$, is fed into MLP $f_\eta$ with $P$ hidden layers to output a SDF value. Both MLP parameters and feature embeddings are optimized through the self-supervised training, more detail is covered in Section \ref{subsec:SematFeature}. 
\end{itemize}

\subsubsection{\textbf{Semantic Features Construction}}
\label{subsec:SematFeature}
To implicitly build a semantic model of scenes, 
our semantic information is obtained from off-the-shelf semantic segmentation or panoptic segmentation algorithms. 


\begin{itemize}
\label{item:snf}
    \item \textbf{Semantic Neural Fields (SNF)}. To increase the robustness of SNF module, in the training stage, we uniformly sample the points along the LiDAR rays, but only the sampled points near the object surface are assigned with semantic labels, the sampled points falling in free space are skipped to obtain the semantic information. In the inferring stage, following the method of geometry feature, each sampled point  $x_i \in \mathbb{R}^3$ retrieves eight semantic feature  embeddings $F \in \mathbb{R}^{H_2}$, from eight corners of corresponding grids through hash table $B_2$ and obtains the $F_k$ of the point via executing tri-linear interpolation. The procedure is repeated to three last level of octree to generate $F_2$, $F_1$ and $F_0$. The concatenated semantic feature embedding, $F_S = F_2 + F_1 +F_0$, is fed into MLP $f_\nu$ with $P$ hidden layers to predict a semantic label. Both MLP parameters and embeddings are optimized through the self-supervised training, our supervision signal is obtained from LiDAR-based semantic segmentation algorithm RangeNet++ \cite{milioto2019rangenet++}. Relied on the feature embedding interpolation and MLPs regression, the SNF is established. The SNF regression method is faster than nearest searching which is shown in Fig. \ref{fig:ente}. 

\begin{figure}[t!]
    \centering

    \includegraphics[width=1\linewidth]{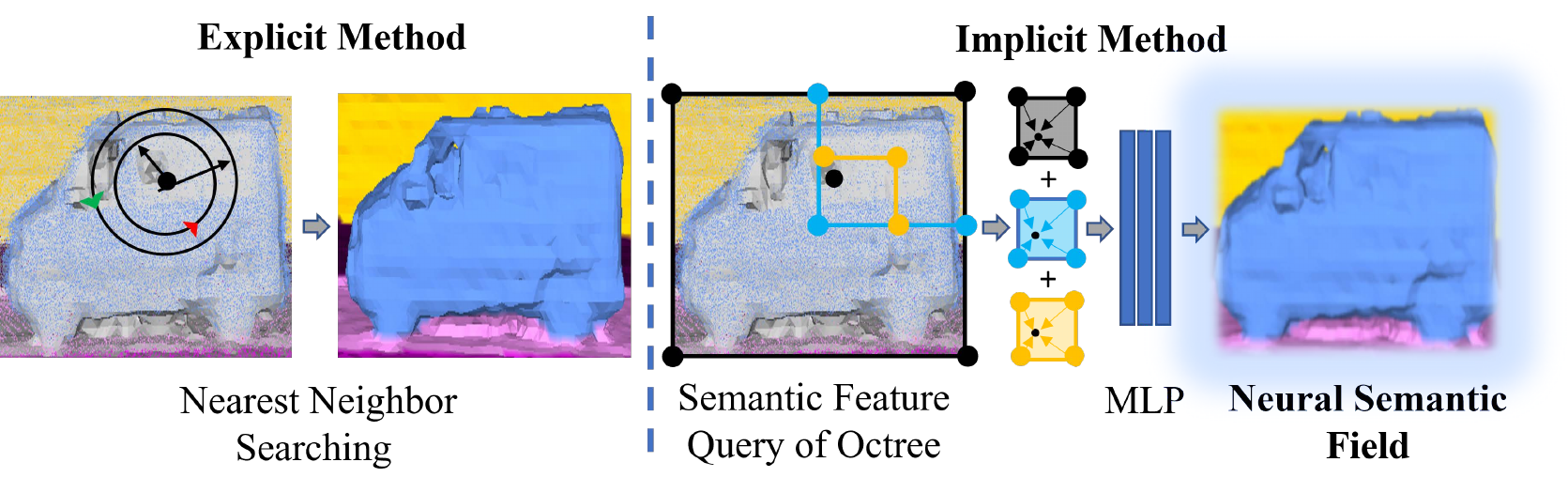}
    
    \caption{ When querying the semantics of a vertex, intuitively, we can assign it with the semantics of the nearest point in the point cloud via time-consuming local searching. The left image illustrates the concept of finding the nearest point to obtain semantics, while the right image shows our implicit SNF method. Our method can achieve faster speed with less memory consumption compared to the nearest point lookup approach.  
    }
    \label{fig:ente}
    \vspace{-0.3cm}
\end{figure}
    \item \textbf{Panoptic Neural Fields (PNF)}. We extend above semantic paradigm to panoptic segmentation. Here, the panoptic feature embedding $F$ not only implicitly contains the category label of things and stuff, but also includes instance IDs of the thing classes. The length of feature embedding $F$ is $H_2$, we then define a ratio $\sigma$ to split the $F$, a part of the feature embedding with length $\sigma H_2$ stores semantic label, the other part of feature embedding with length $(1 - \sigma) H_2$ represents the instance IDs of the thing. The consistent instance ID is obtained from off-the-shelf video panoptic segmentation algorithms. After we concatenate the sum-up feature embedding $F_S$, $F_S$ is also split into two parts by following ratio $\sigma$. Next, we respectively feed the two uneven parts of feature embeddings into two separate MLPs, to regress final panoptic labels.

\end{itemize}
\vspace{-0.1cm}
\subsection{Training and Loss Function}
\label{subsec:TrainLoss}
As mentioned above, LiDAR is able to provide accurate range measurements. Thus, the true SDF value is directly utilized to supervise the training, we picked up the distance from sampled point to the beam endpoint as supervision signal. For semantic labels and instance labels, the output of semantic segmentation or panoptic segmentation is selected as supervision signal to form a pseudo-supervision paradigm. 

We uniformly sample $N$ points along the LiDAR rays to train SDF MLP and semantic MLPs, half of the points sampled near the objects surface and the other half sampled within the free space. For SDF value, binary cross entropy is exploited for the loss function, $L_1$. Given a sampled point $x_i \in \mathbb{R}^3$ and signed distance to the surface $d_i$, $L_1$ is denoted as:
\begin{equation}
    L_1=\frac{log(1/(1+e^{\frac{f_\eta(x_i)}{\alpha}}))}{1/(1+e^{d_i/\alpha})} + \frac{log(1-1/(1+e^{\frac{f_\eta(x_i)}{\alpha}}))}{1-1/(1+e^{d_i/\alpha})}
\end{equation}
where $f_\eta(x_i)$ represents the SDF value of geometry MLP output, $\alpha$ is a hyper-parameter. Following iSDF \cite{ortiz2022isdf}, we apply Eikonal regularization to add another term, $L_2$, into loss function:
\begin{equation}
    L_2=| \|\bigtriangledown_x f(x; \eta) \|-1|
\end{equation}

For incremental mapping, there is a forgetting issue when the map size increases. Similar to \cite{zhong2023shine}, we add a regularization term, $L_3$, to the loss function:
\begin{equation}
    L_3=\sum_{i\in N_{all}}\beta_i(\eta^t_i-\eta^{t-1}_i)^2
\end{equation}
where $N_{all}$ refers to all points in this scan. $\eta^t_i$ stands for the MLP parameter of current iteration, $\eta^{t-1}_i$ means the parameter of history iteration. $\beta_i$ is defined as a importance weight:
\begin{equation}
    \beta_i=min(\beta^{t-1}_i + \sum\|\bigtriangledown_{\eta_i} L_1(x_i;d_i) \|,\beta_m)
\end{equation}
where $\beta^{t-1}_i$ refers to the previous importance weight, $\beta_m $ is a constant value to prevent gradient explosion. 

For training stage of the semantic labels, we only enroll the points sampled near the objects' surface. Given sampled point $x_i$ and its semantic label $s_i$, we leverage multi-class cross entropy as loss function :
\begin{equation}
    L_4=\sum^{c-1}_{i=0}s_ilog(S(f_\nu(x_i)))
\end{equation}
where $c$ is the number of semantic categories. $f_\nu(x_i)$ represents the output of the semantic MLP. $S$ is Softmax function. 

Then, we treat instance ID prediction as a multi-class task as well, our panoptic loss function is:
\begin{equation}
    L_5=\sum^{q-1}_{i=0}y_ilog(S(f_\mu(x_i)))
\end{equation}
where $q$ represents the number of instance. $y_i$ refers to supervision ID, which is generated from an off-the-shelf panoptic segmentation algorithm. $f_\mu(x_i)$ is the output of instance MLP. 

Our mapping method provides batch mode and incremental mode. The complete loss function is designed as follow based on the different mapping modes.
\begin{itemize}
    \item Loss of incremental semantic mapping:
\begin{equation}
    L_{is}=L_1 + \lambda_2 L_2 +\lambda_3 L_3+\lambda_4 L_4
\end{equation}
    \item Loss of incremental panoptic mapping:
    \begin{equation}
    L_{ip}=L_1 + \lambda_2 L_2 +\lambda_3 L_3+\lambda_4 L_4 +\lambda_5 L_5
\end{equation}
    \item Loss of batch-based semantic mapping:
\begin{equation}
    L_{bs}=L_1 + \lambda_2 L_2 +\lambda_4 L_4
\end{equation}
    \item Loss of batch-based panoptic mapping:
\begin{equation}
    L_{bp}=L_1 + \lambda_2 L_2 +\lambda_4 L_4 +\lambda_5 L_5
\end{equation}
\end{itemize}
where $\lambda_2, \lambda_3, \lambda_4, \lambda_5$ are hyper-parameters used to adjust the weight of loss function.
\begin{figure*}[h]

    \centering
\captionsetup{position=top}
\captionsetup[subfigure]{labelformat=empty}
\subfloat[ GT Point Cloud  ]{\includegraphics[width=.24\linewidth]{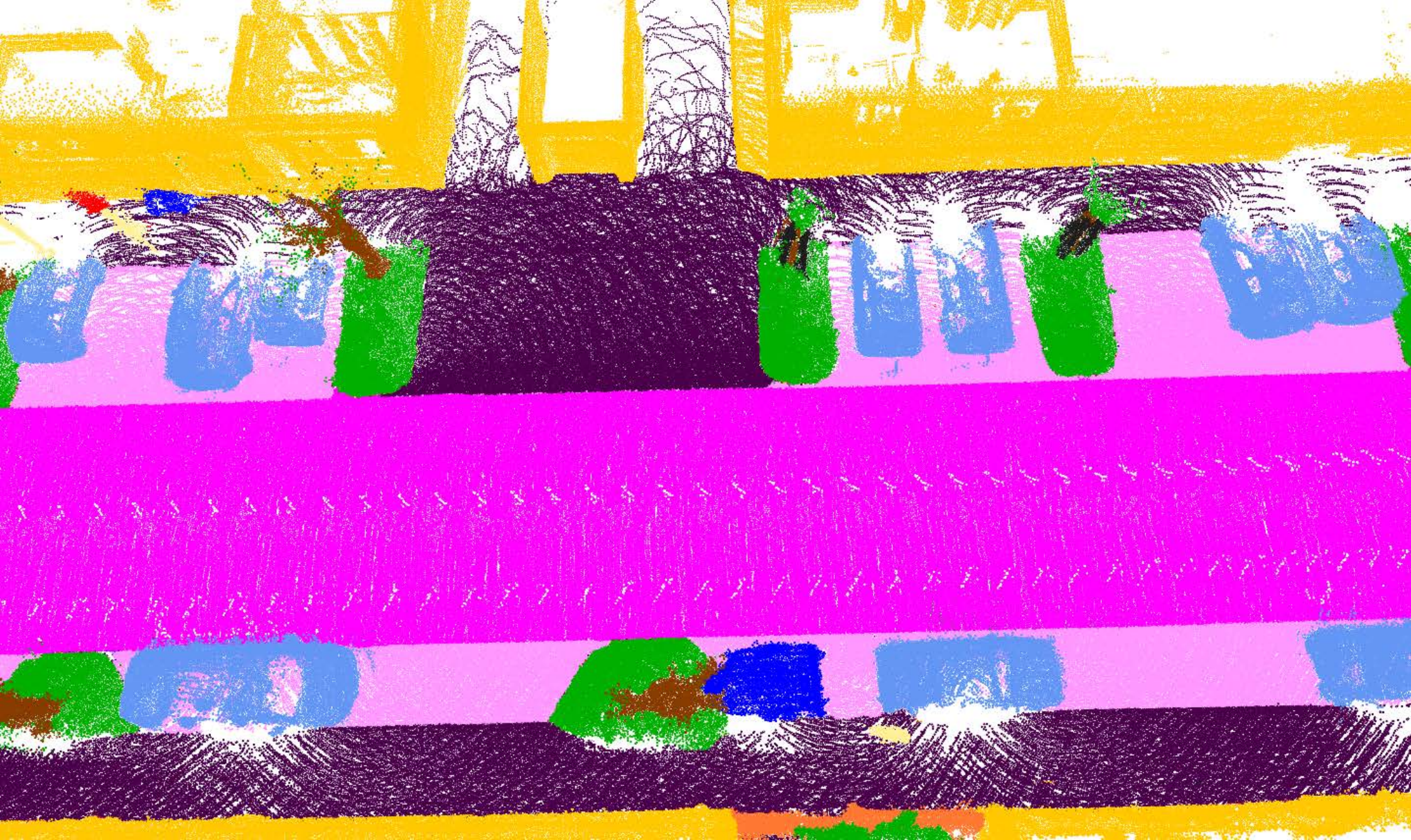}}
\hspace{0.05em} 
\vspace{-0.15em}
\subfloat[Suma++ \cite{behley2018efficient}]{\includegraphics[width=.24\linewidth]{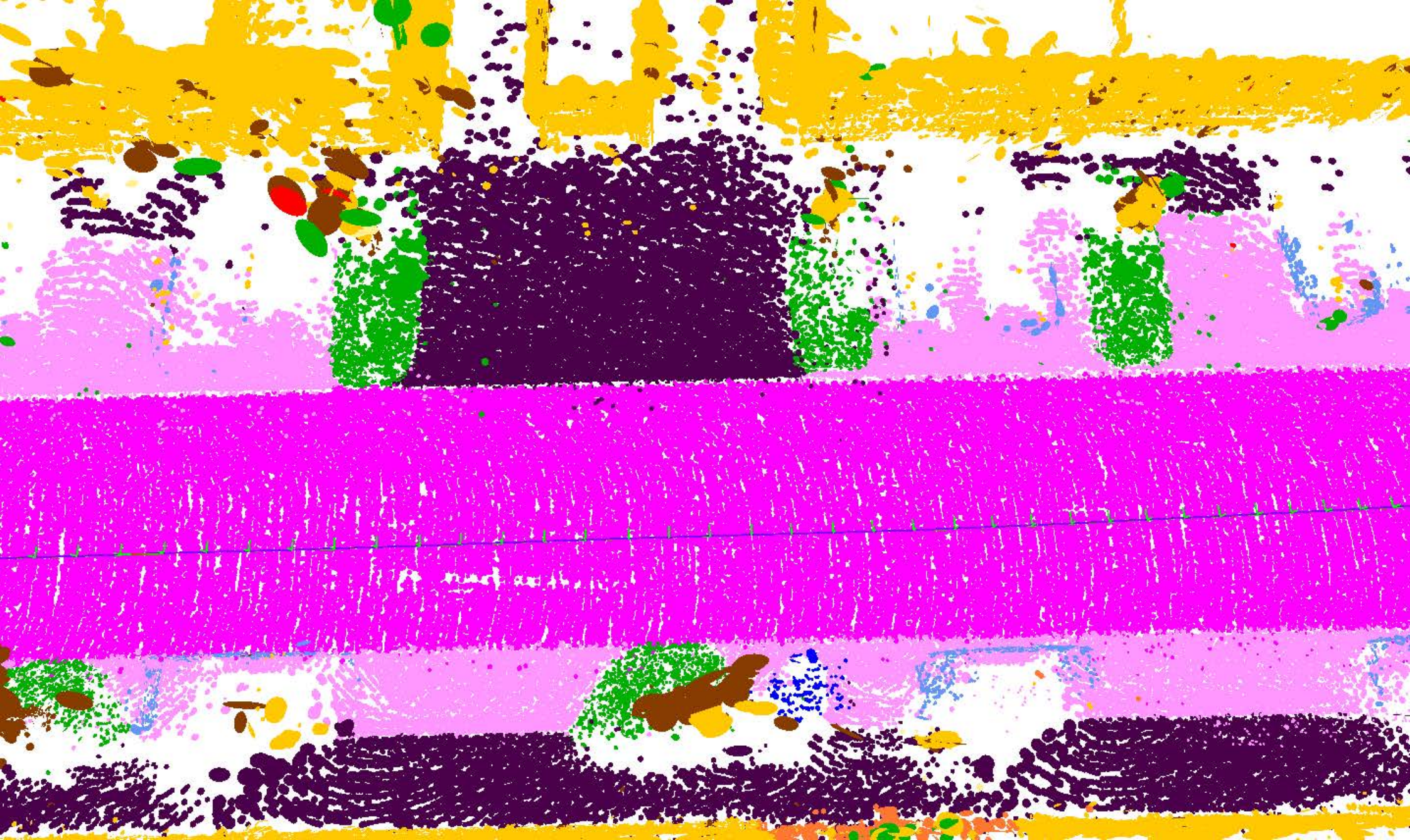}}
\hspace{0.05em} 
\vspace{-0.15em}
\subfloat[LODE \cite{li2023lode}]{\includegraphics[width=.24\linewidth]{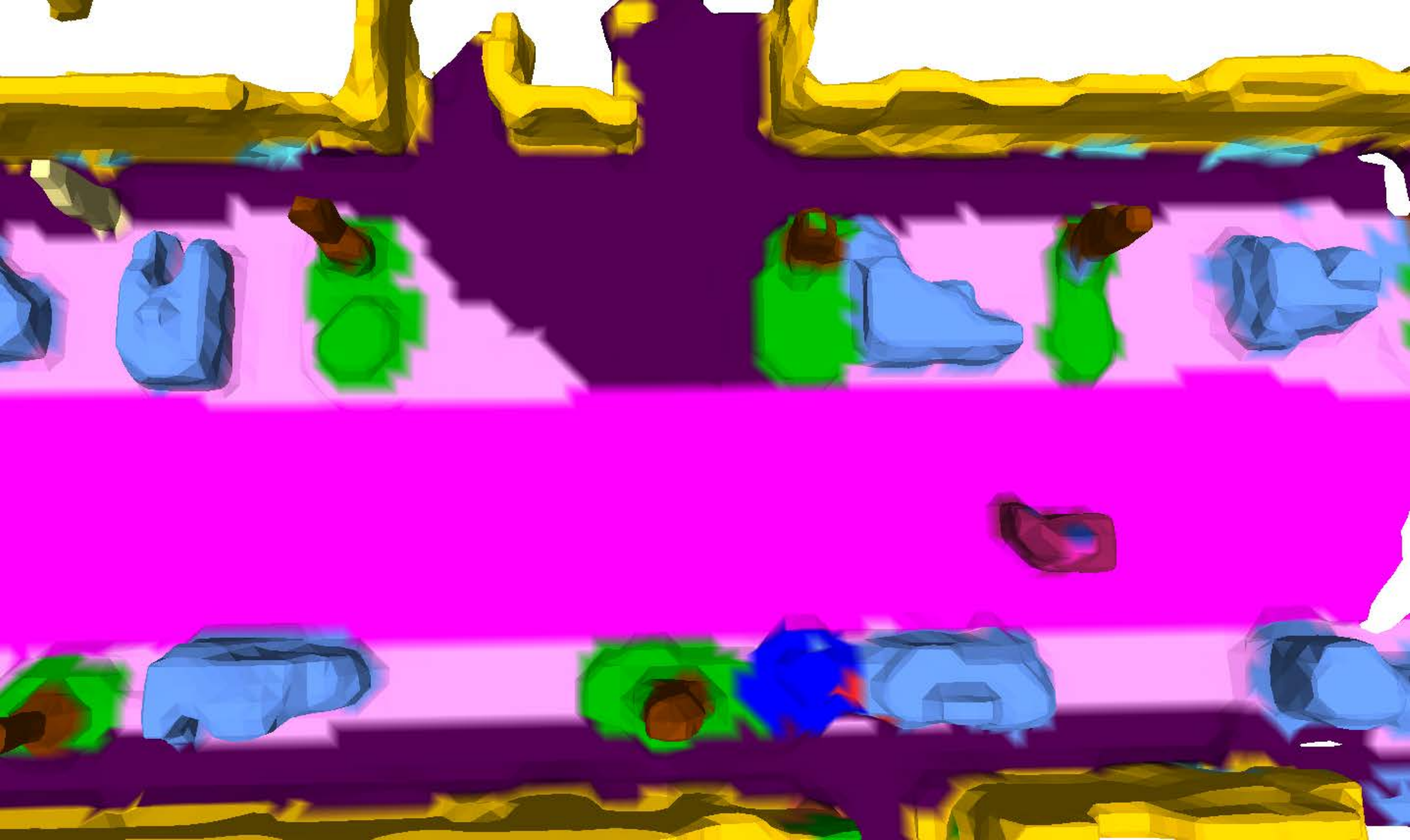}}
\hspace{0.05em} 
\vspace{-0.15em}
\subfloat[Ours]{\includegraphics[width=.24\linewidth]{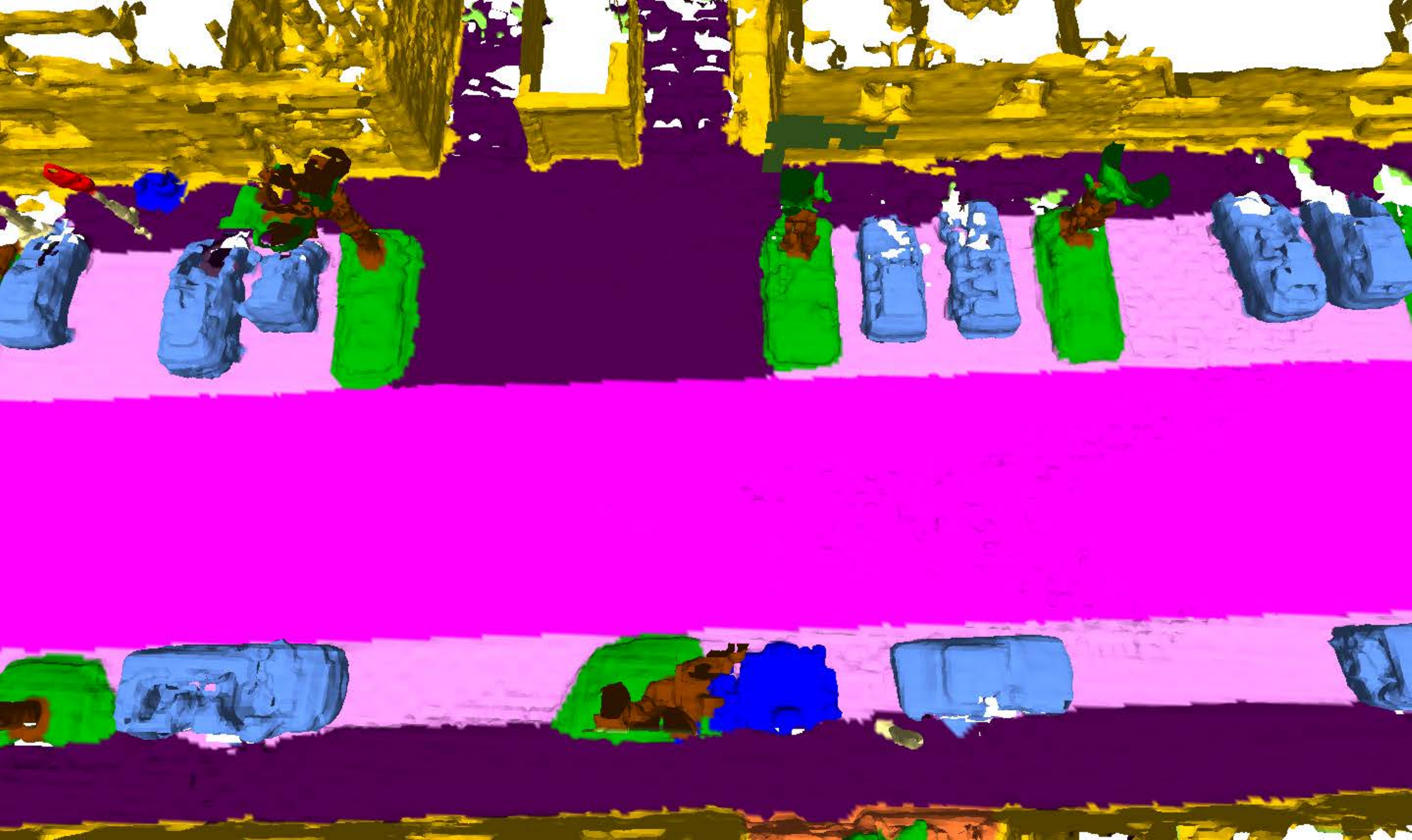}}
\vspace{-0.15em}

\subfloat{\includegraphics[width=.24\linewidth]{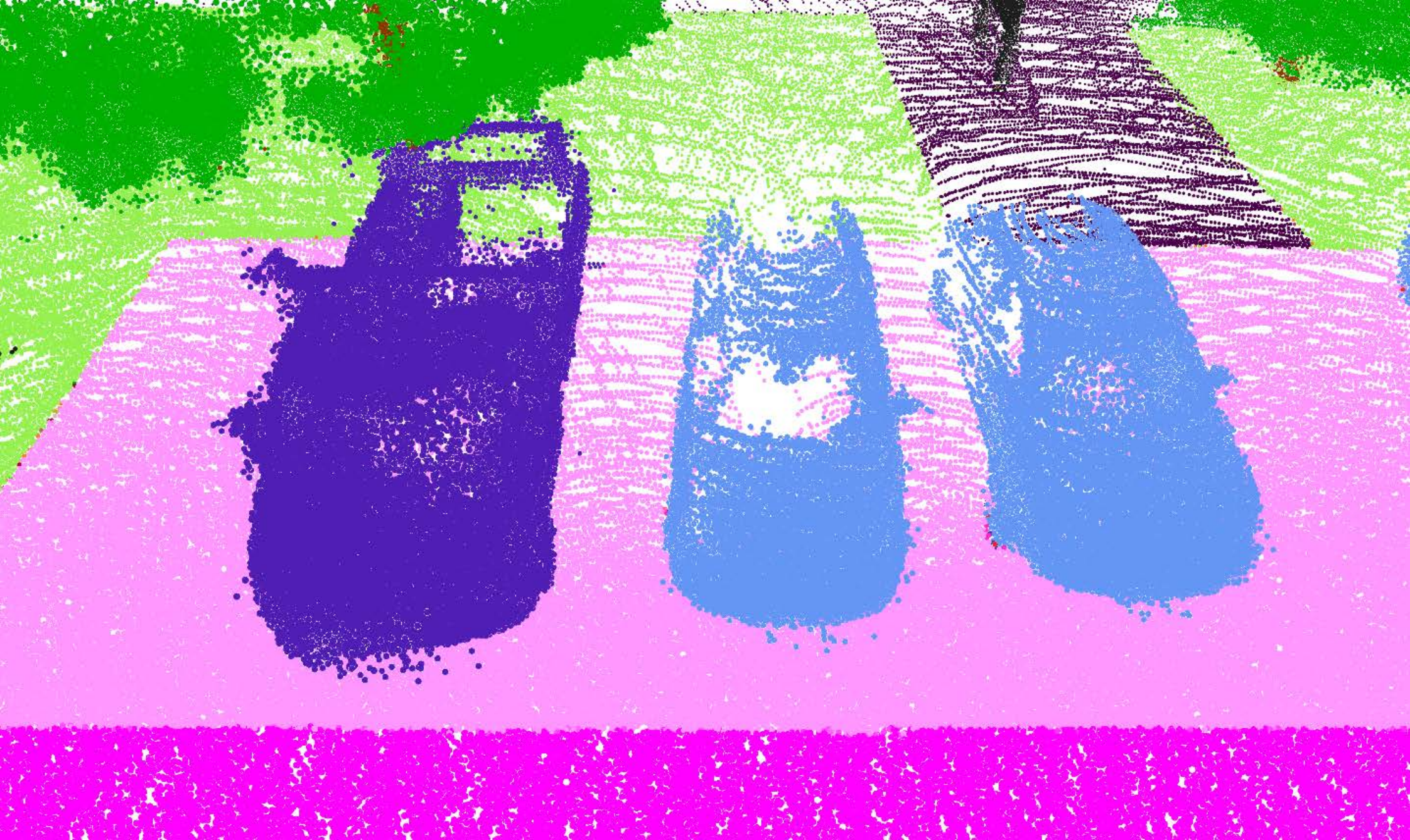}}
\hspace{0.05em} 
\vspace{-0.3em}
\subfloat{\includegraphics[width=.24\linewidth]{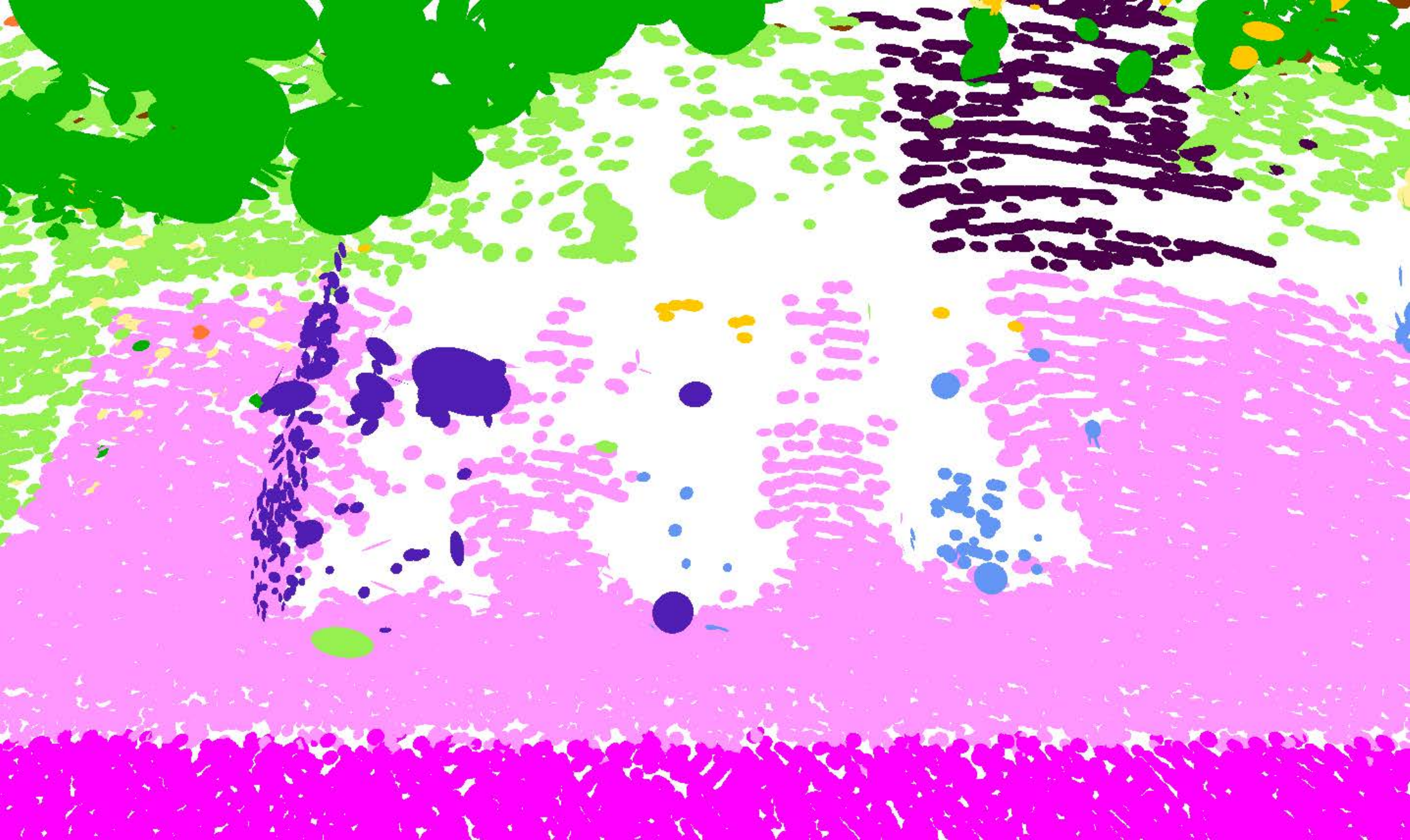}} 
\hspace{0.05em} 
\vspace{-0.3em}
\subfloat{\includegraphics[width=.24\linewidth]{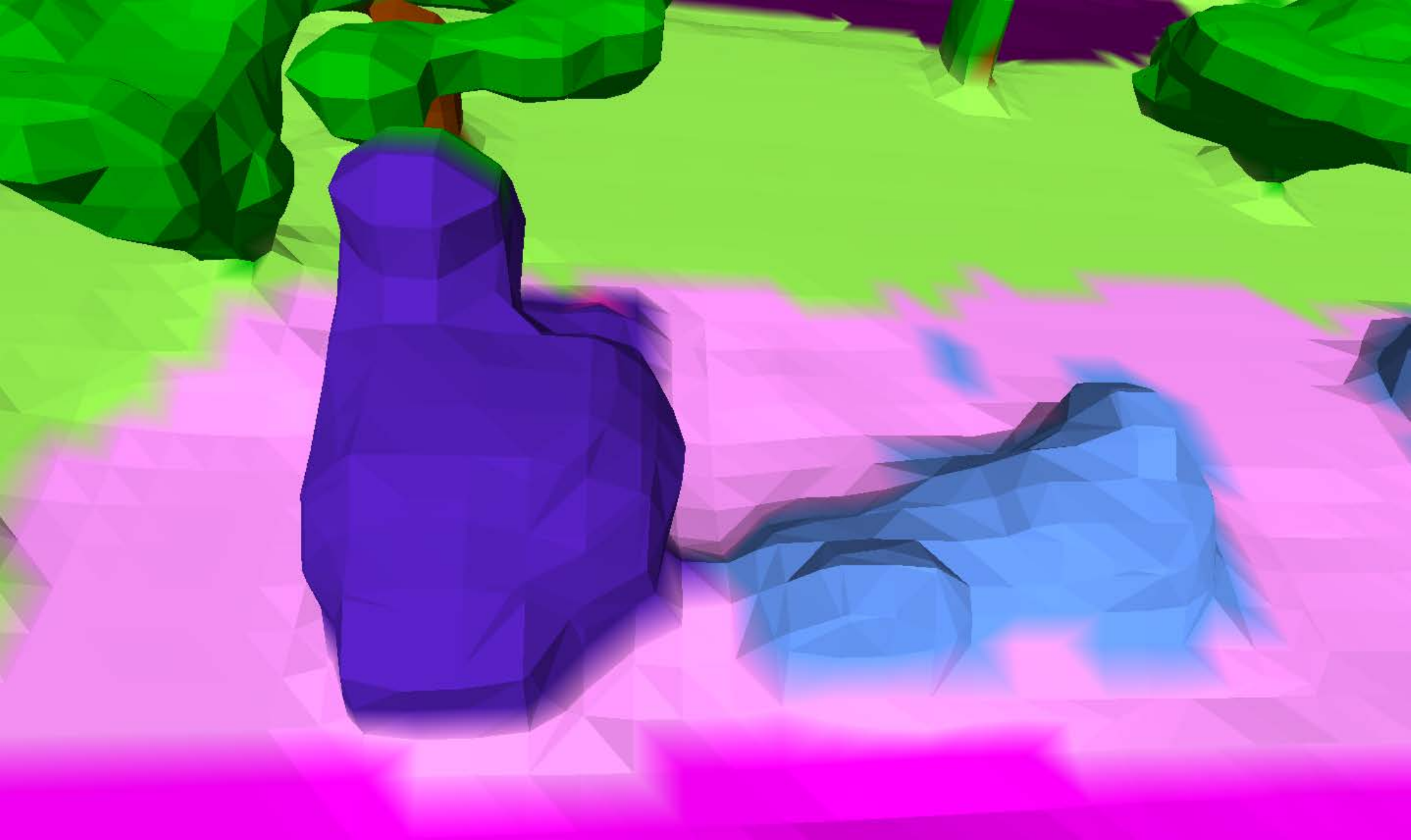}} 
\hspace{0.05em} 
\vspace{-0.3em}
\subfloat{\includegraphics[width=.24\linewidth]{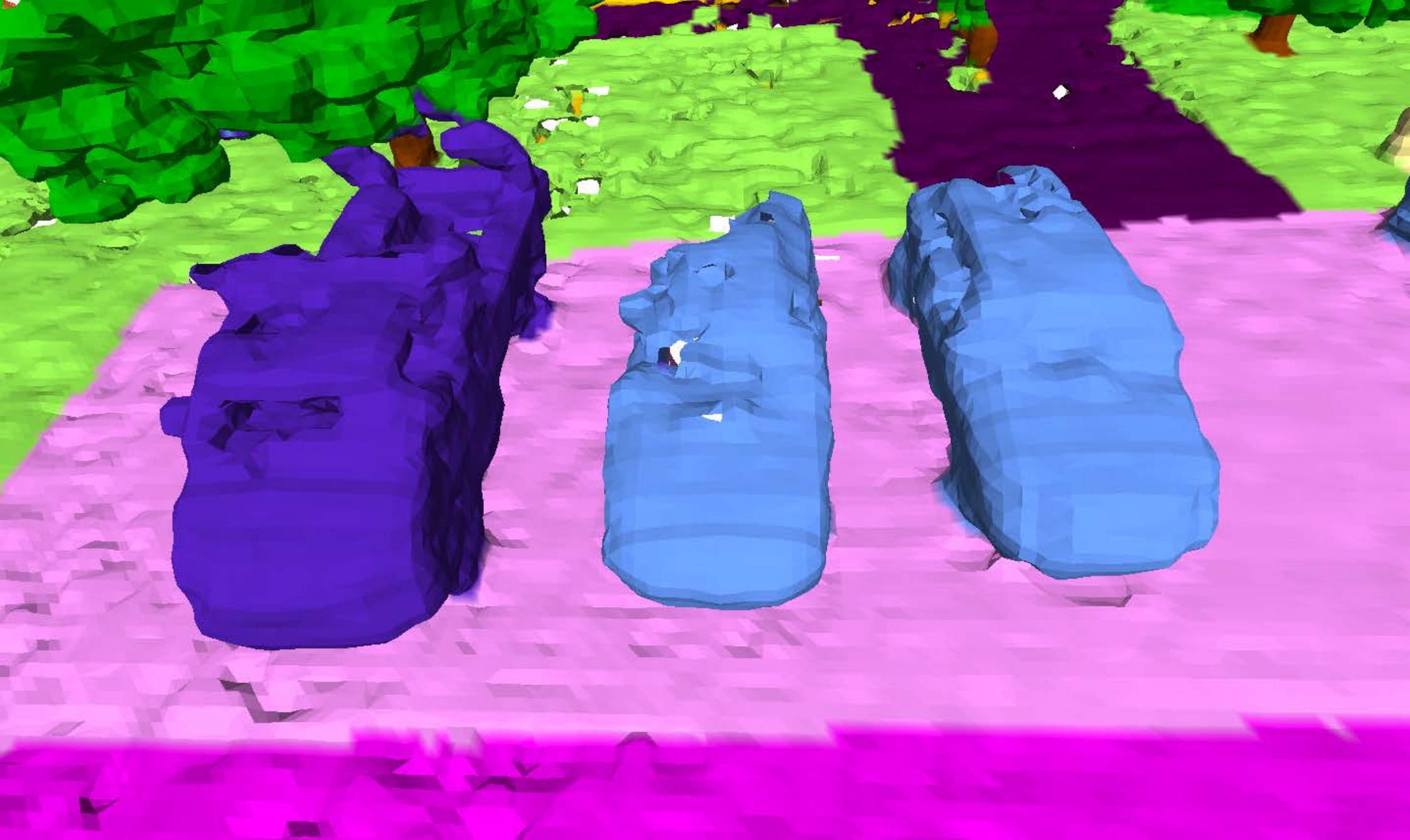}}
\vspace{-0.3em}
\subfloat{\includegraphics[width=.98\linewidth]{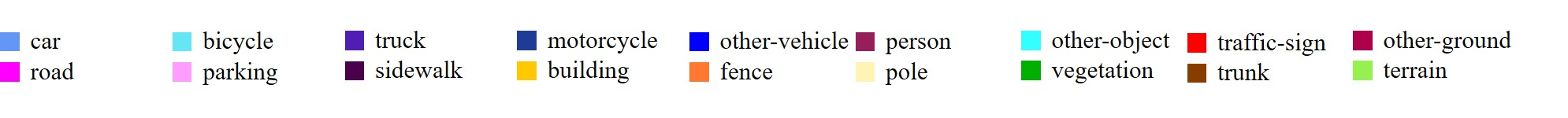}}
\vspace{-0.4cm}
\caption{The visualization of our 3D semantic reconstruction results on SemanticKITTI via the comparison with other related methods ("GT"= Ground Truth). The mapping results in first row are tested on SemanticKitti 00 sequence, and the second row is tested on SemanticKitti 05 sequence. Our map is denser compared to Suma++, more precise and complete in terms of semantics compared to LODE. Best viewed with zoom in.}
\vspace{-0.8cm}
\label{fig:different-scene}
\end{figure*}
\subsection{Map Merge Strategy}
\label{subsec:MapMerge}
Considering the limitations of device memory, it is often impractical to input all the data at once to construct an entire large-scale map, especially  the ray casting of NeRF-based methods for urban-level mapping amplifies this issue. Instead, our solution incrementally feed data in batches to create submaps, which are finally merged to form a complete map. Furthermore, data collection for large-scale areas is typically distributed across multiple agents. Therefore, an off-the-shelf LiDAR odometry can be adopted to estimate the global poses of submaps, and final map is integrated by calculating the overlapping correlation of submaps. 



\begin{figure*}[h]

    \centering
\captionsetup{position=top}
\captionsetup[subfigure]{labelformat=empty}
\subfloat[ SK-02]{\includegraphics[width=.24\linewidth]{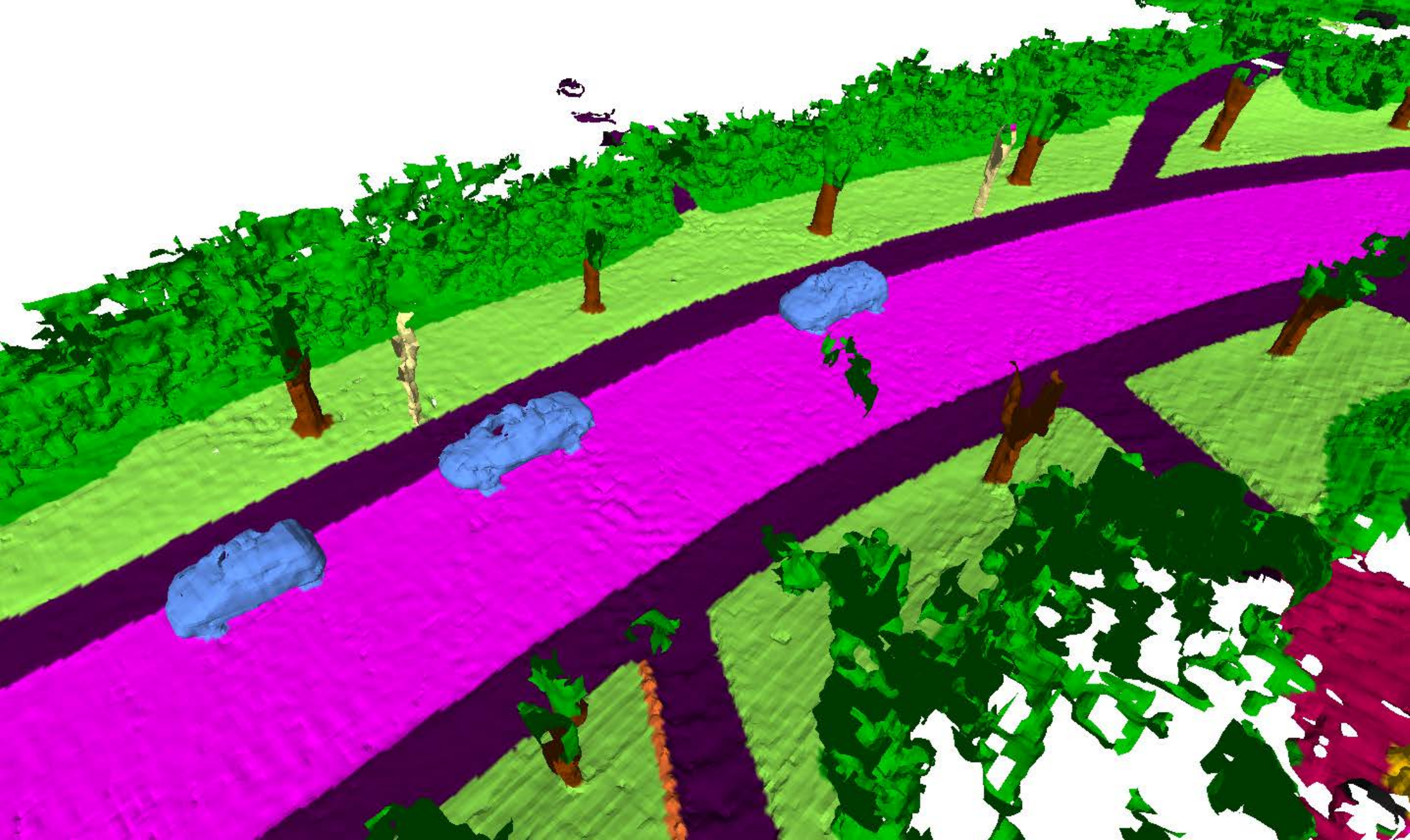}}
\vspace{-0.1em}
\vspace{-0.1em}
\subfloat[SK-05]{\includegraphics[width=.24\linewidth]{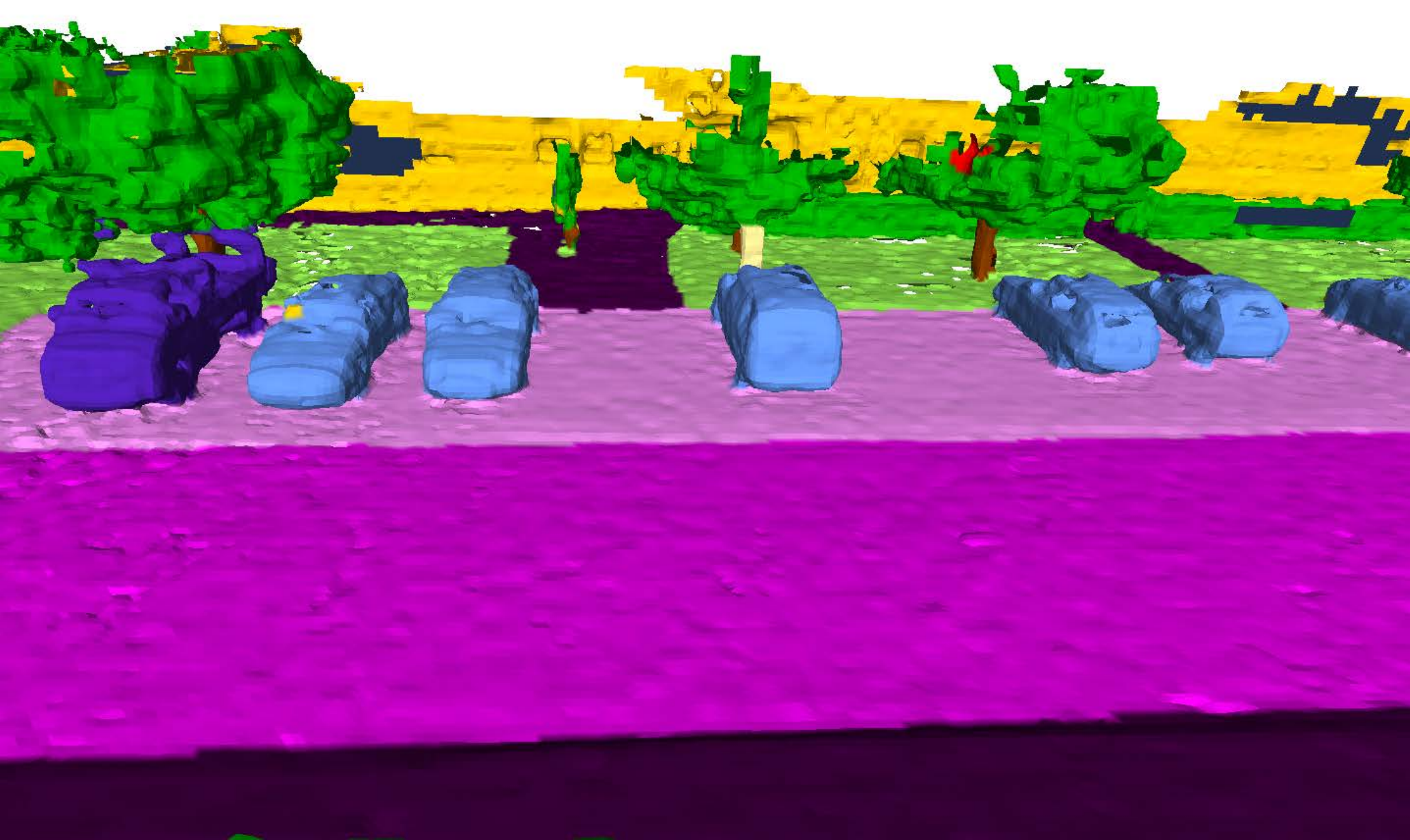}}
\vspace{-0.1em}
\subfloat[SK-03]{\includegraphics[width=.24\linewidth]{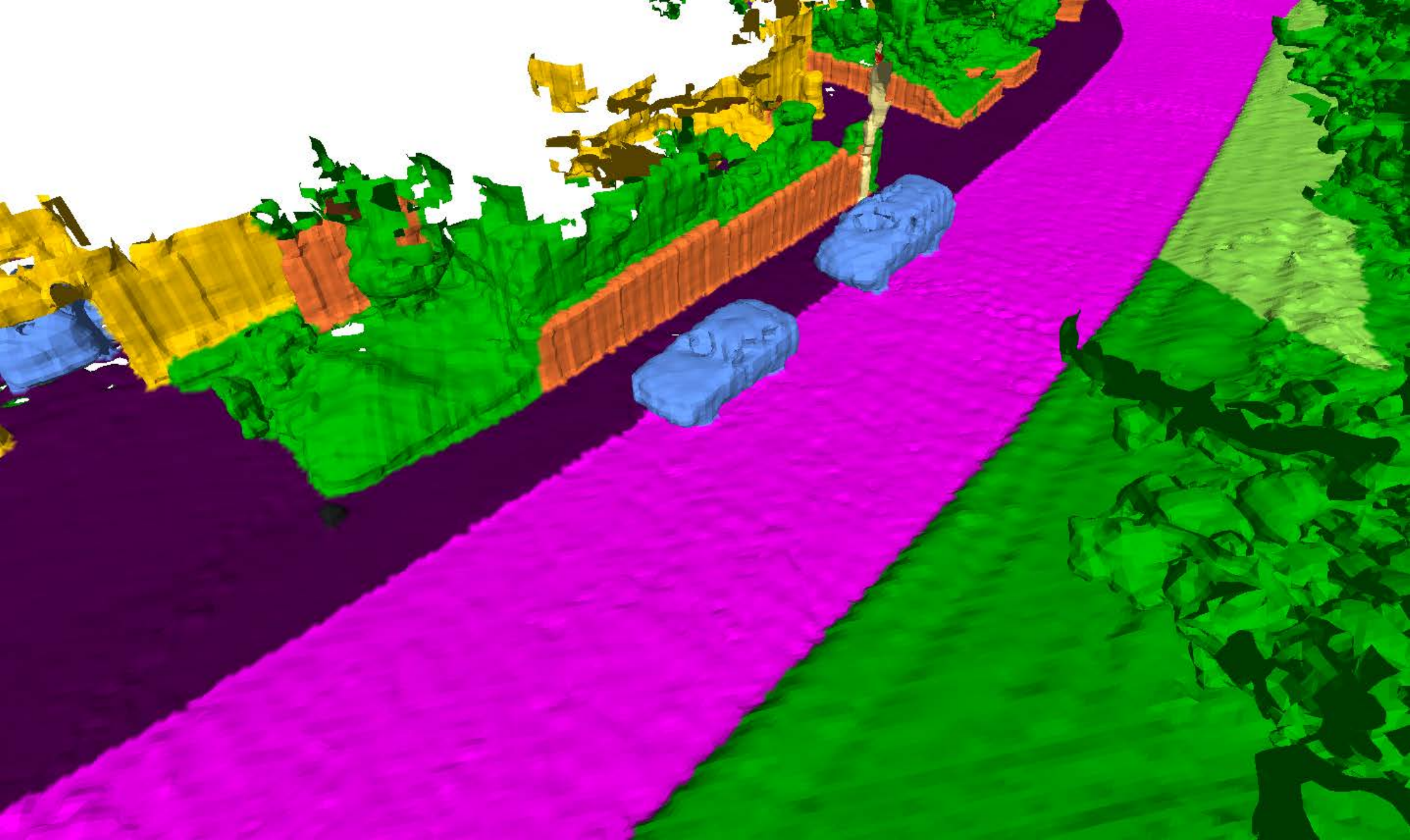}}
\vspace{-0.1em}
\subfloat[SK-07]{\includegraphics[width=.24\linewidth]{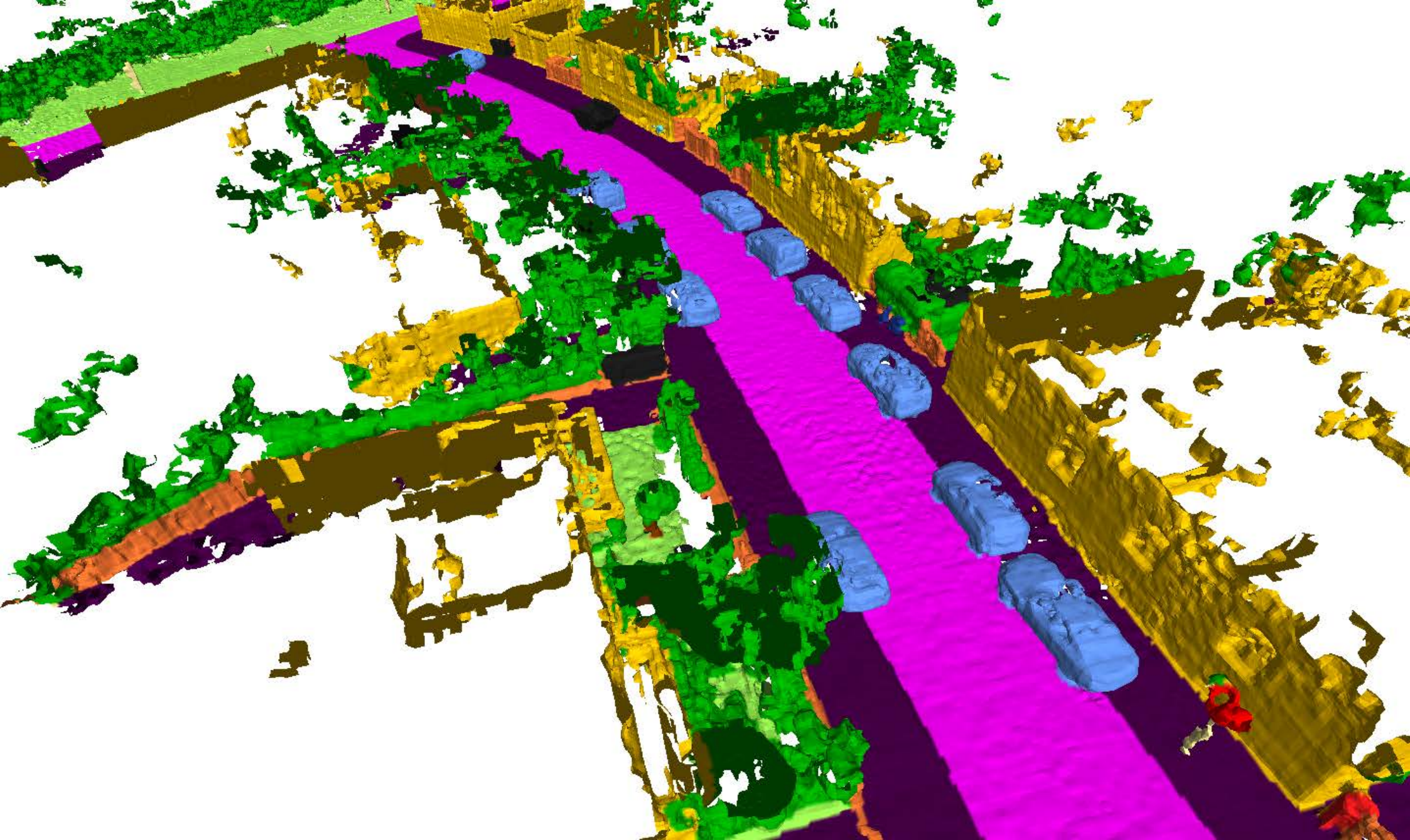}}

\subfloat{\includegraphics[width=.24\linewidth]{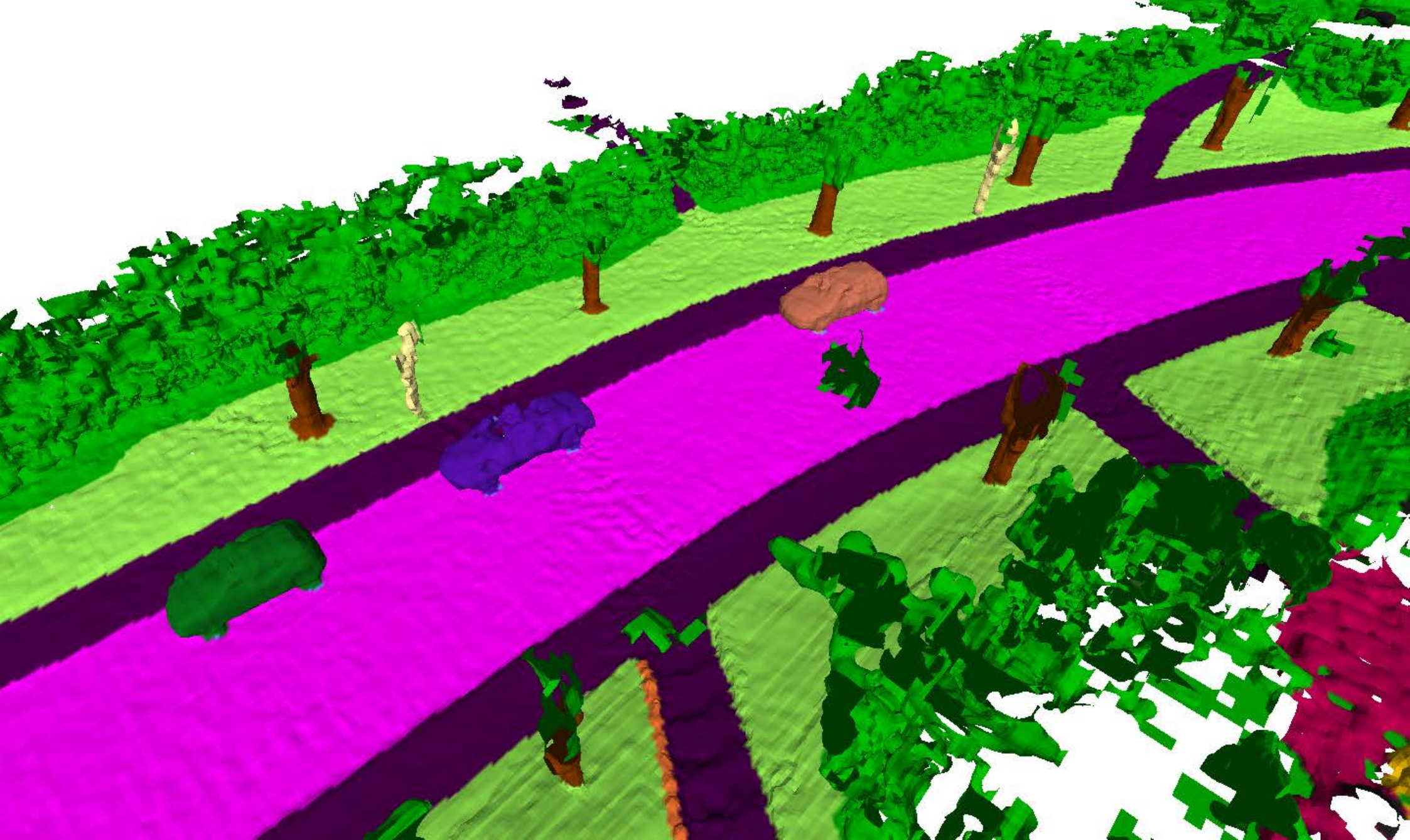}}
\vspace{-0.1em}
\subfloat{\includegraphics[width=.24\linewidth]{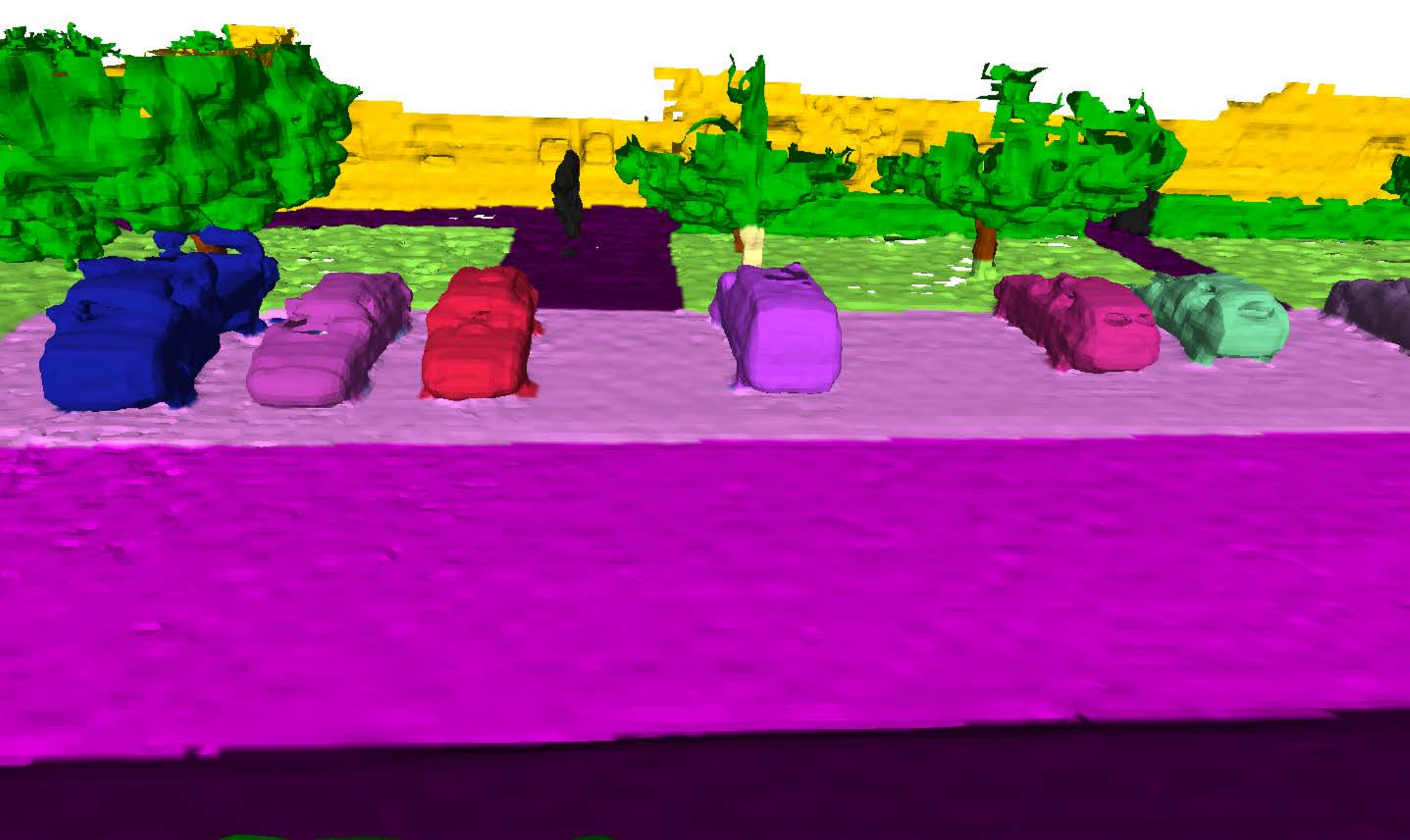}} 
\vspace{-0.1em}
\subfloat{\includegraphics[width=.24\linewidth]{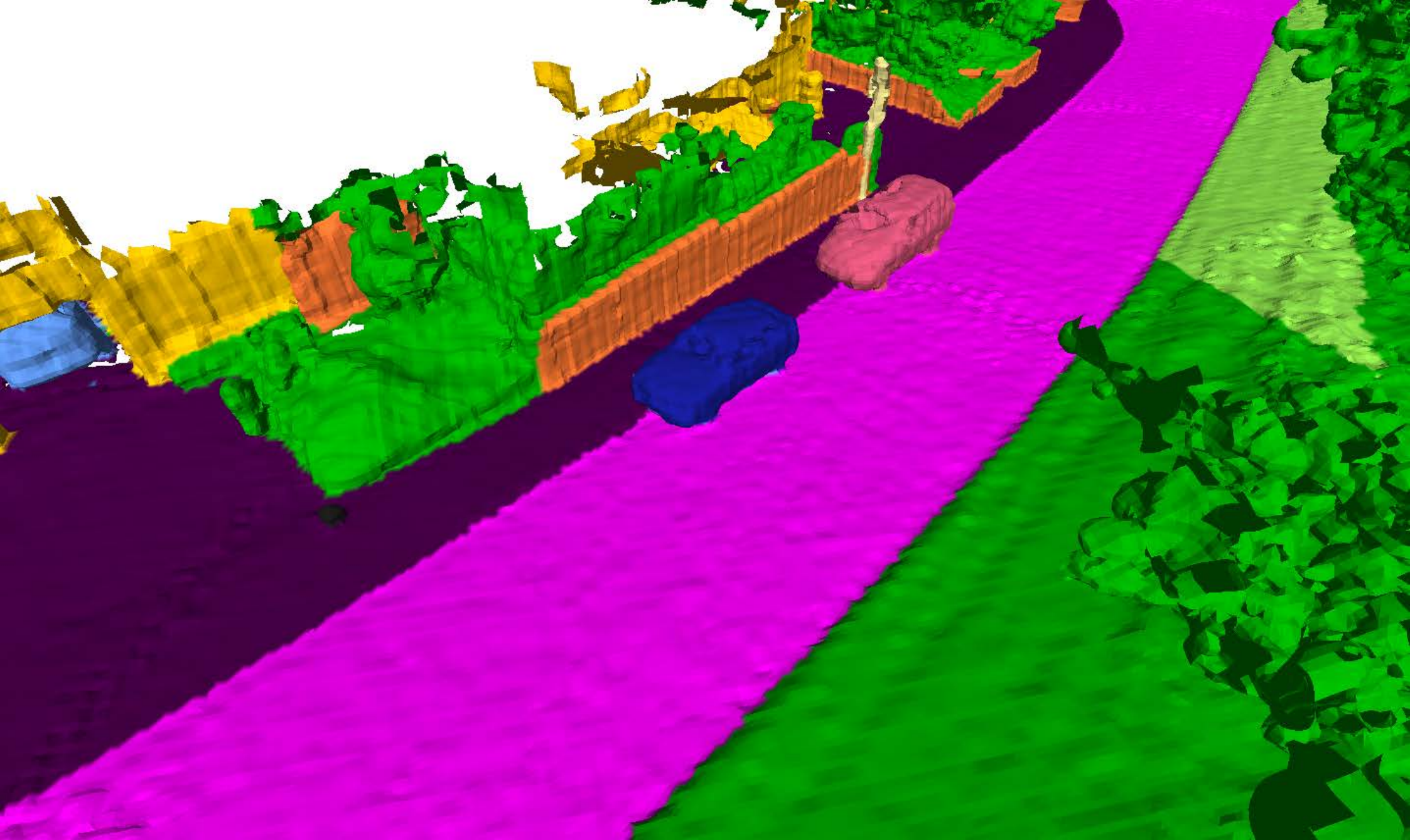}} 
\vspace{-0.1em}
\subfloat{\includegraphics[width=.24\linewidth]{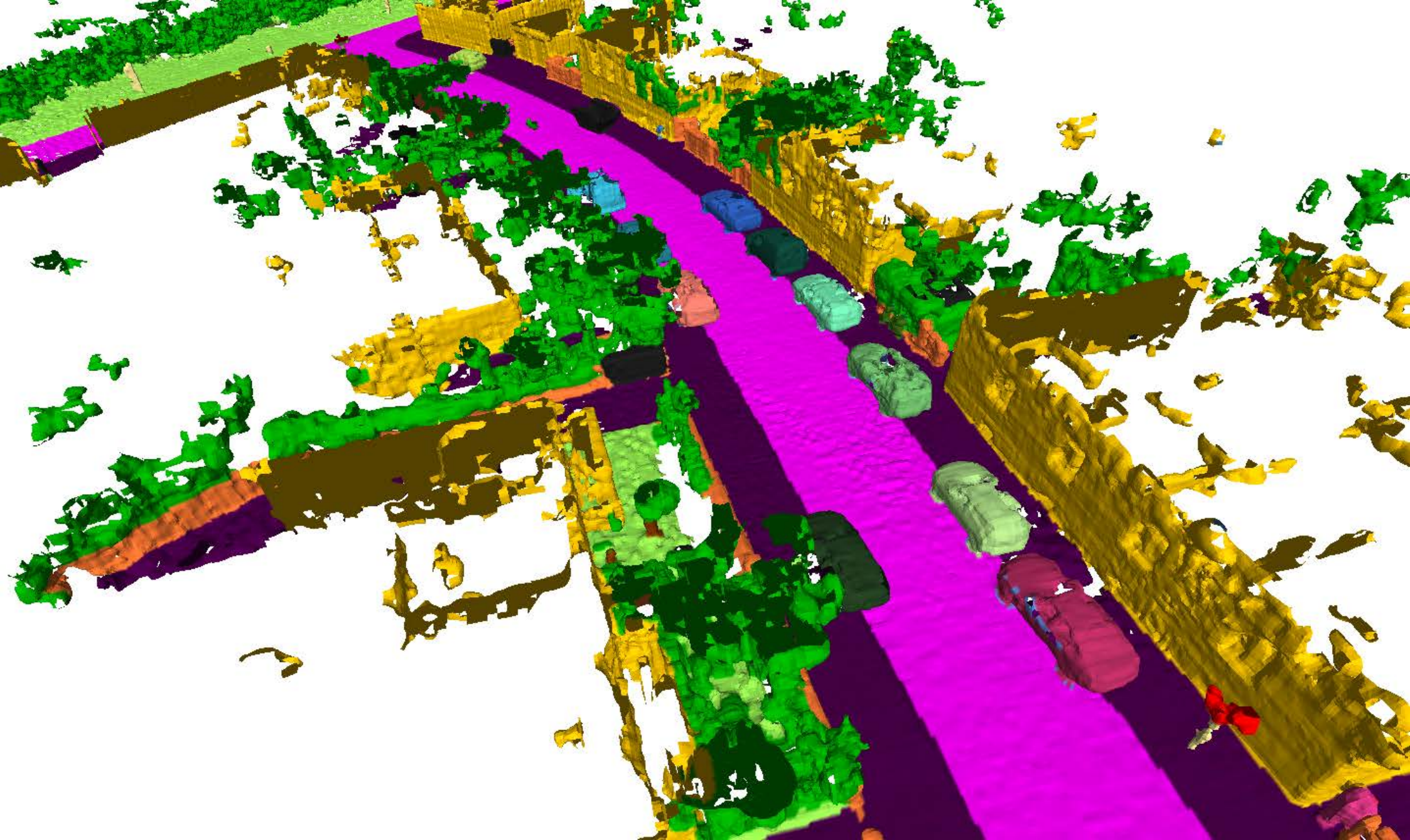}}

\subfloat[ SP-00]{\includegraphics[width=.24\linewidth]{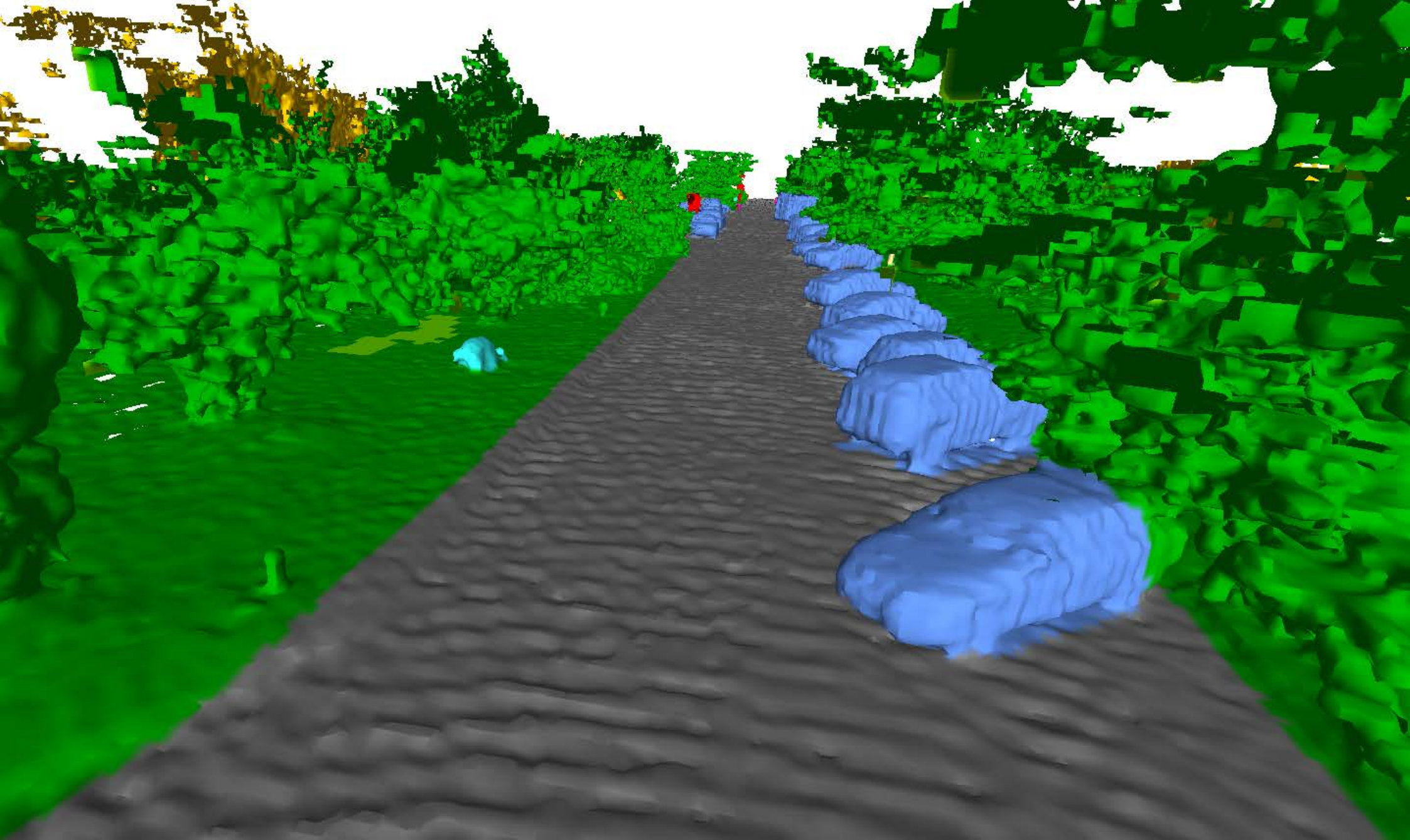}}
\vspace{-0.1em}
\vspace{-0.1em}
\subfloat[SP-01]{\includegraphics[width=.24\linewidth]{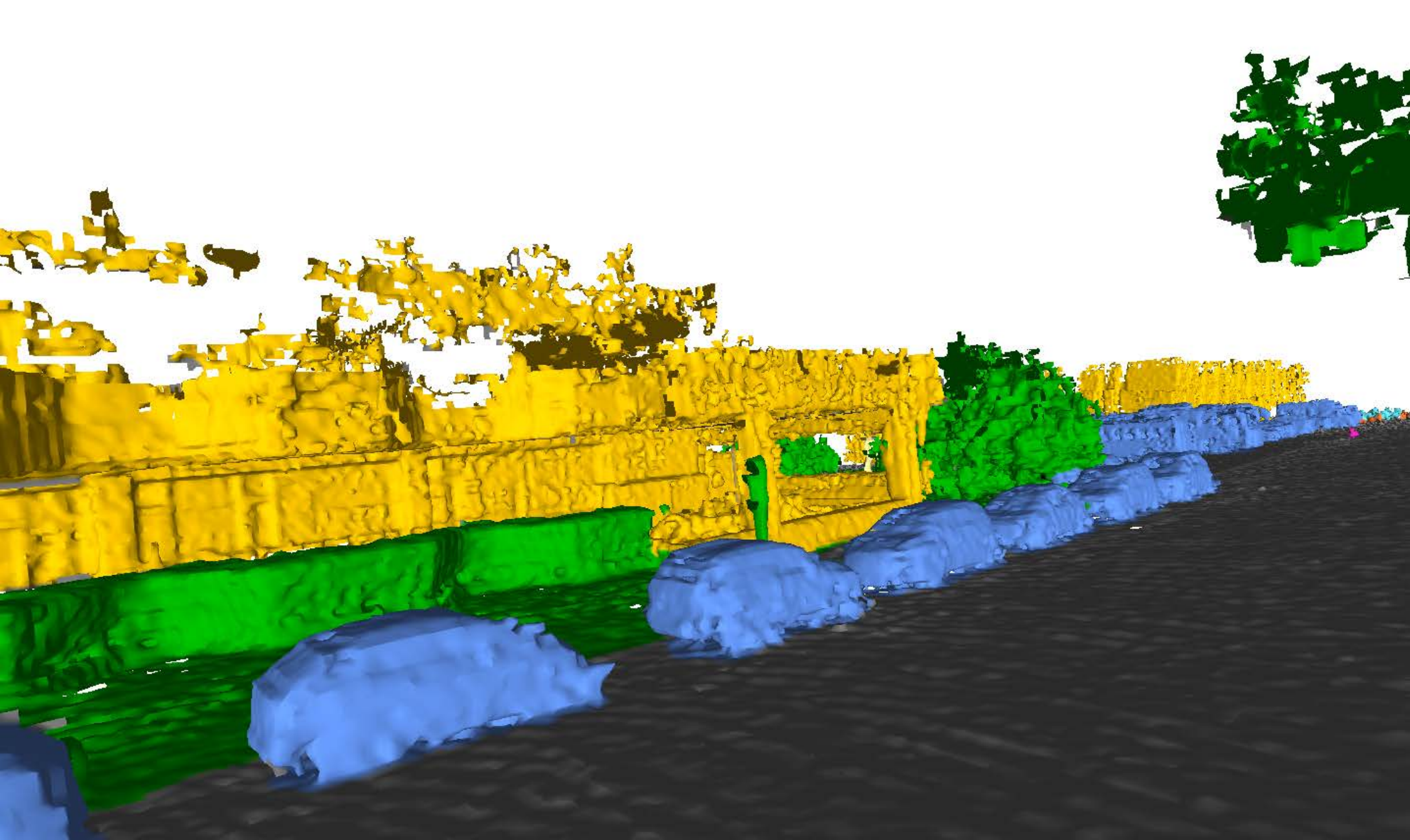}}
\vspace{-0.1em}
\subfloat[SP-04]{\includegraphics[width=.24\linewidth]{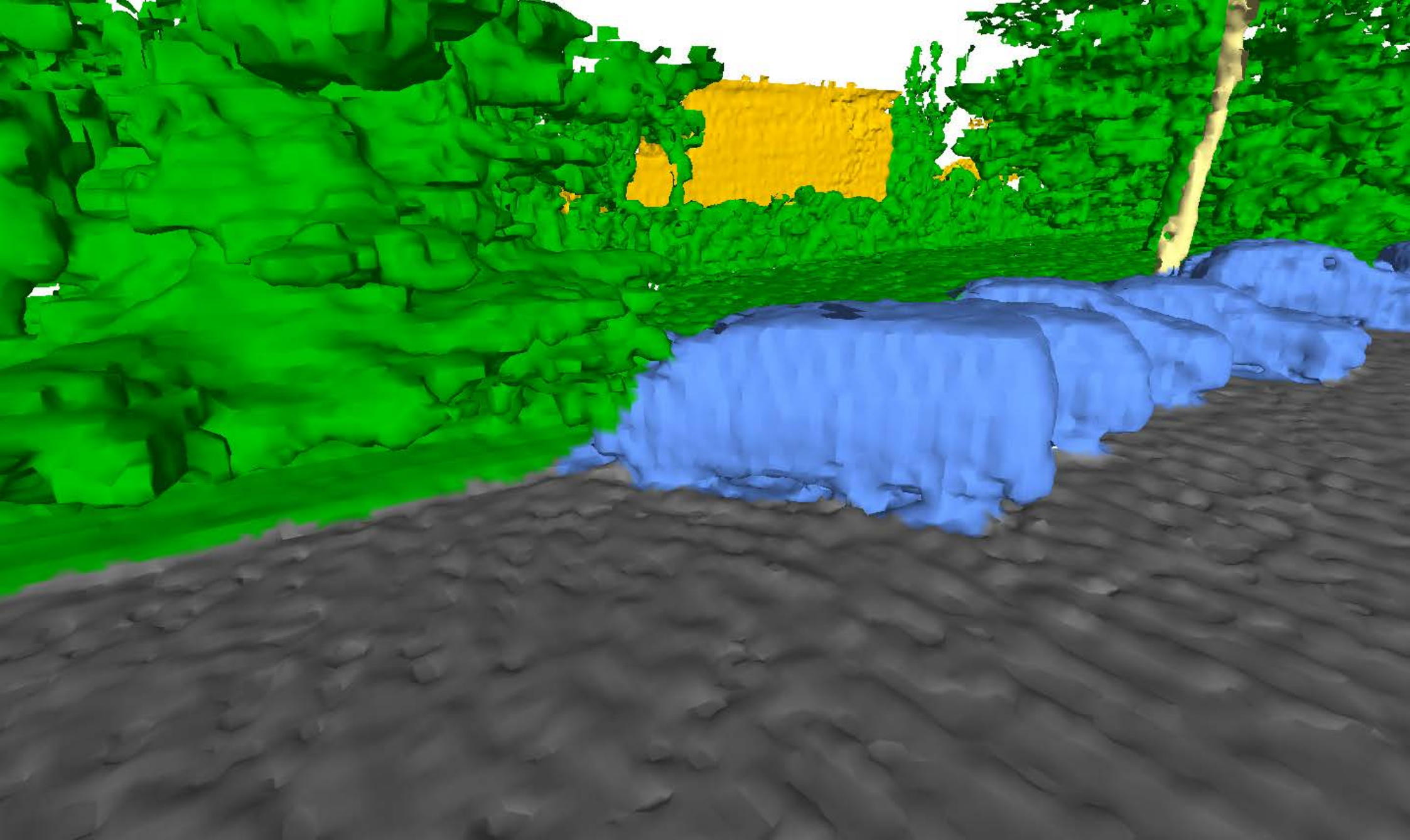}}
\vspace{-0.1em}
\subfloat[SP-05]{\includegraphics[width=.24\linewidth]{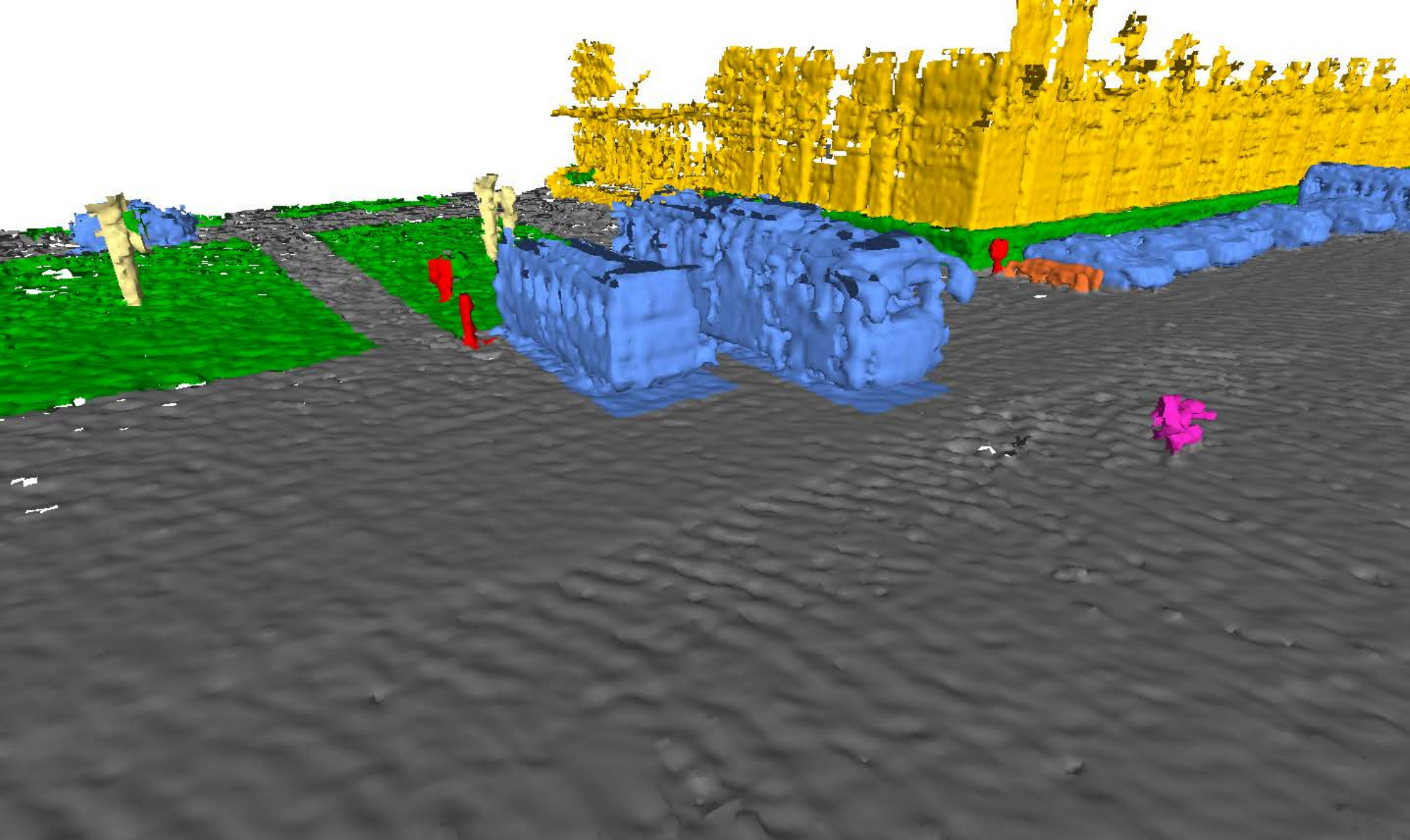}}

\subfloat{\includegraphics[width=.24\linewidth]{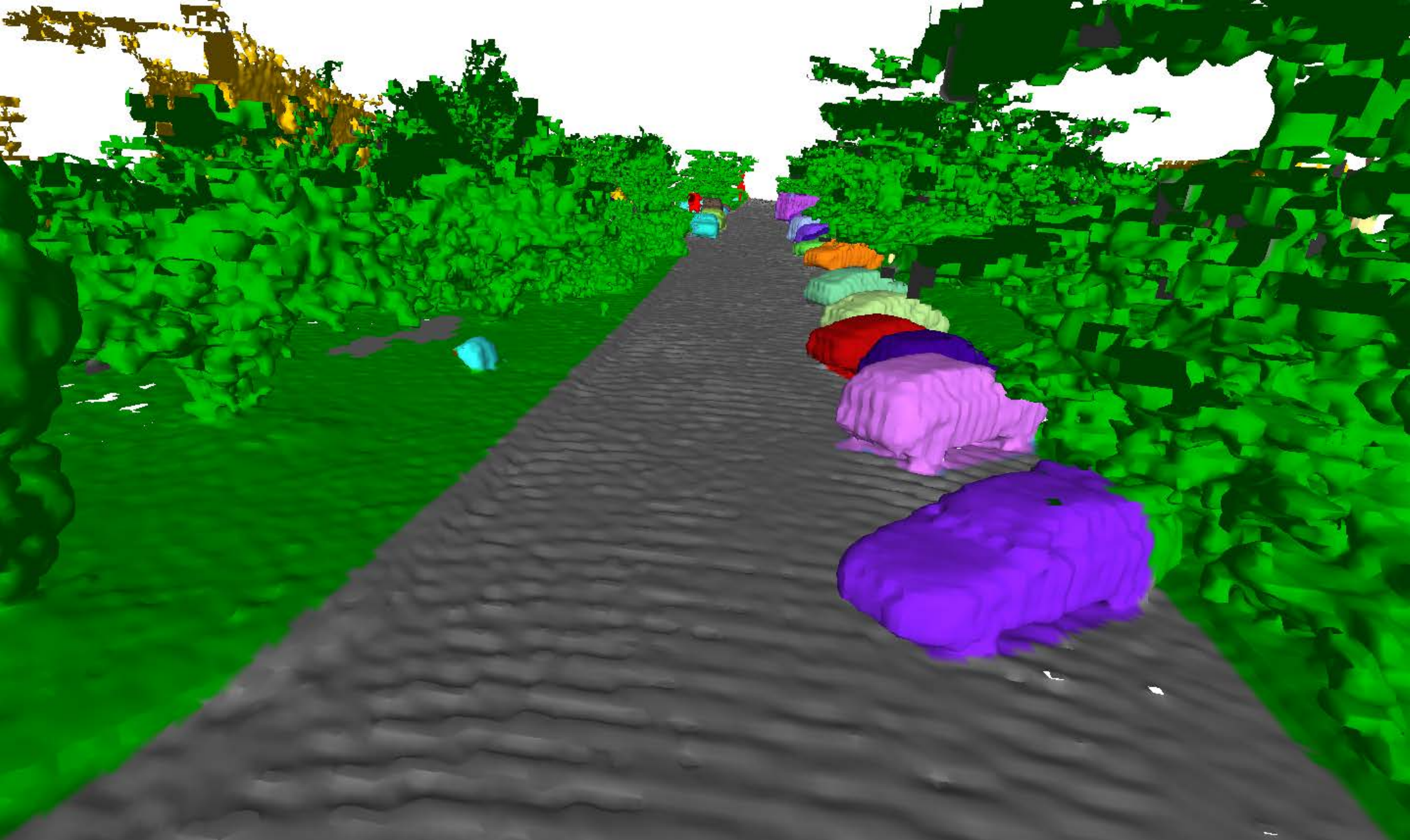}}
\vspace{-0.1em}
\subfloat{\includegraphics[width=.24\linewidth]{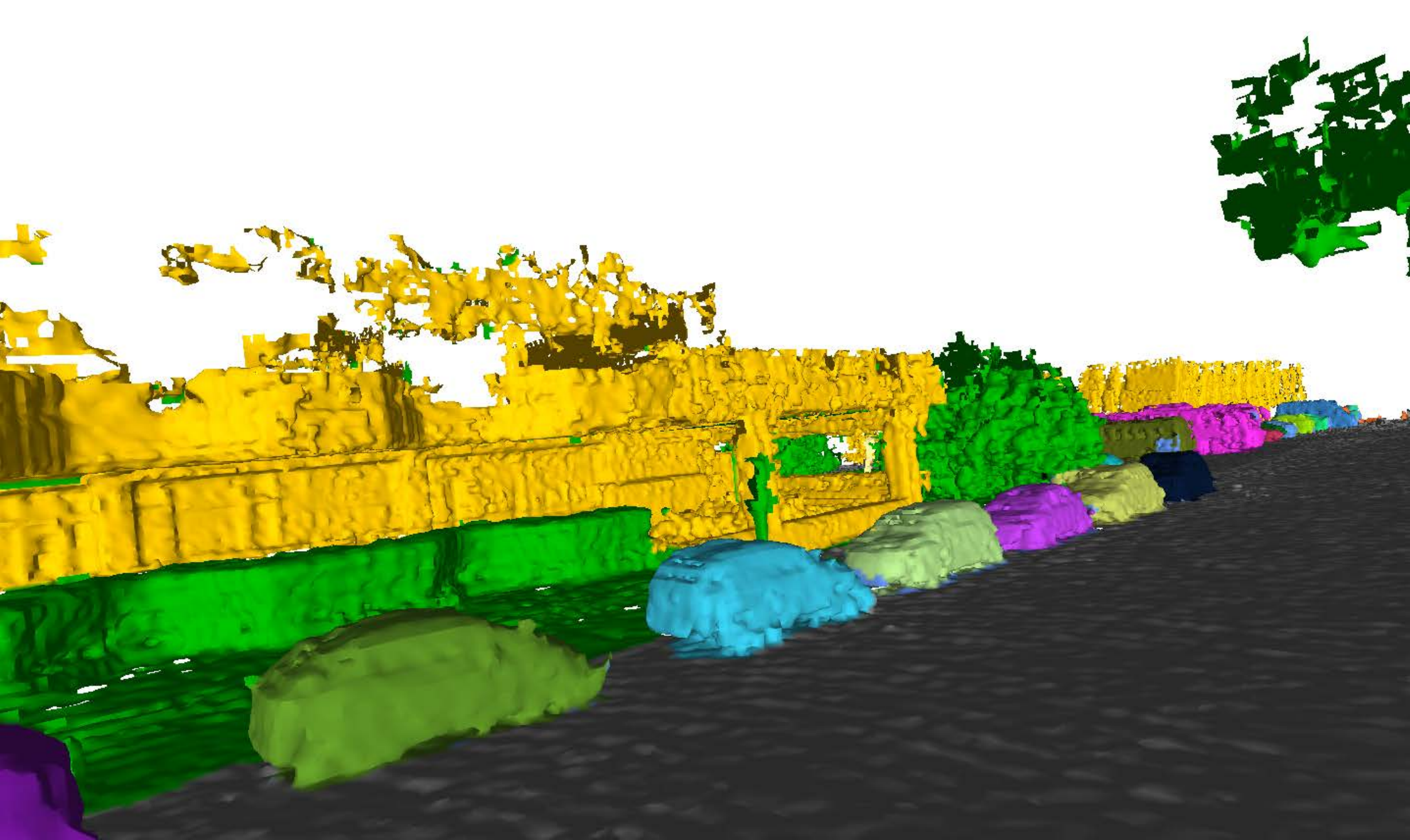}} 
\vspace{-0.1em}
\subfloat{\includegraphics[width=.24\linewidth]{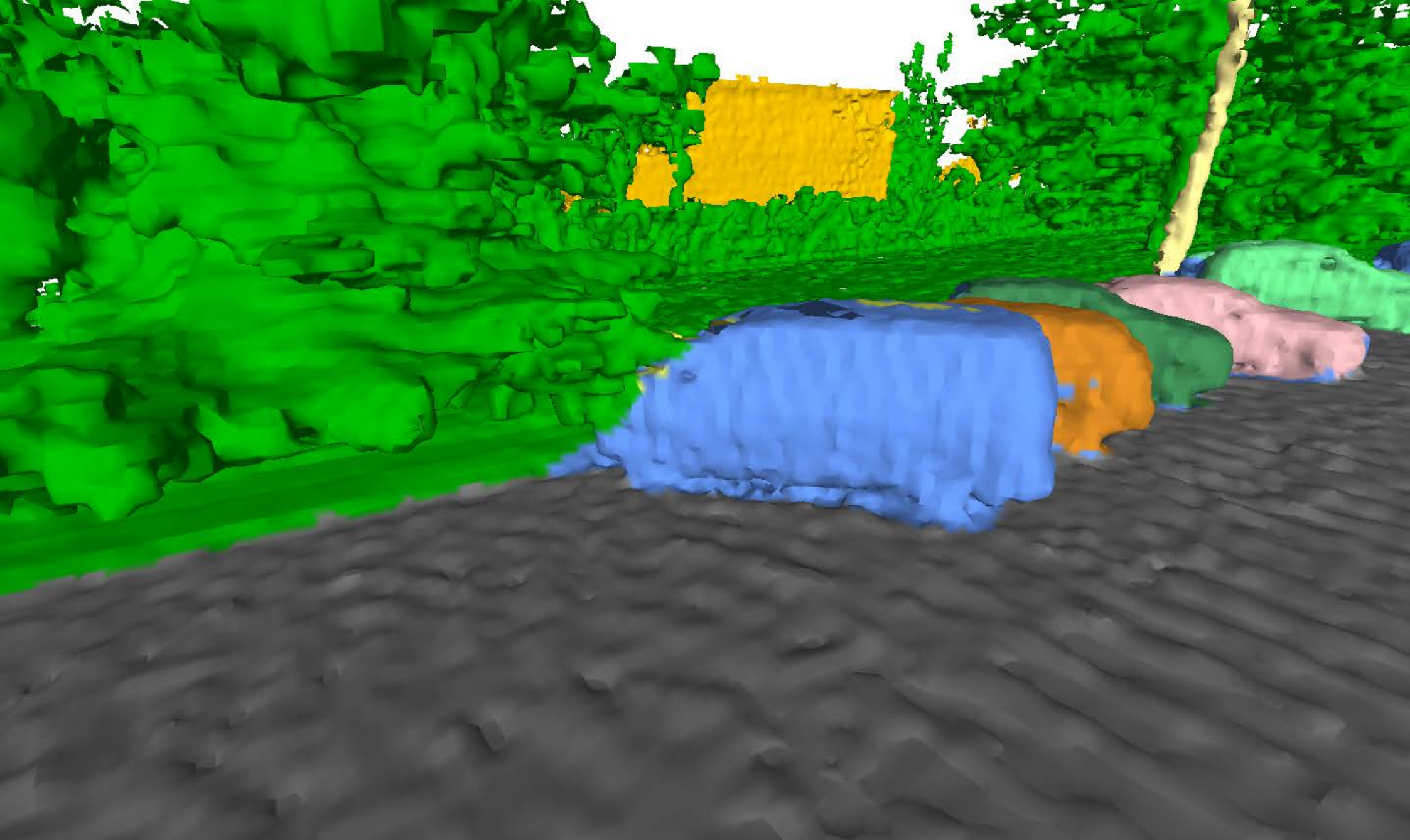}} 
\vspace{-0.1em}
\subfloat{\includegraphics[width=.24\linewidth]{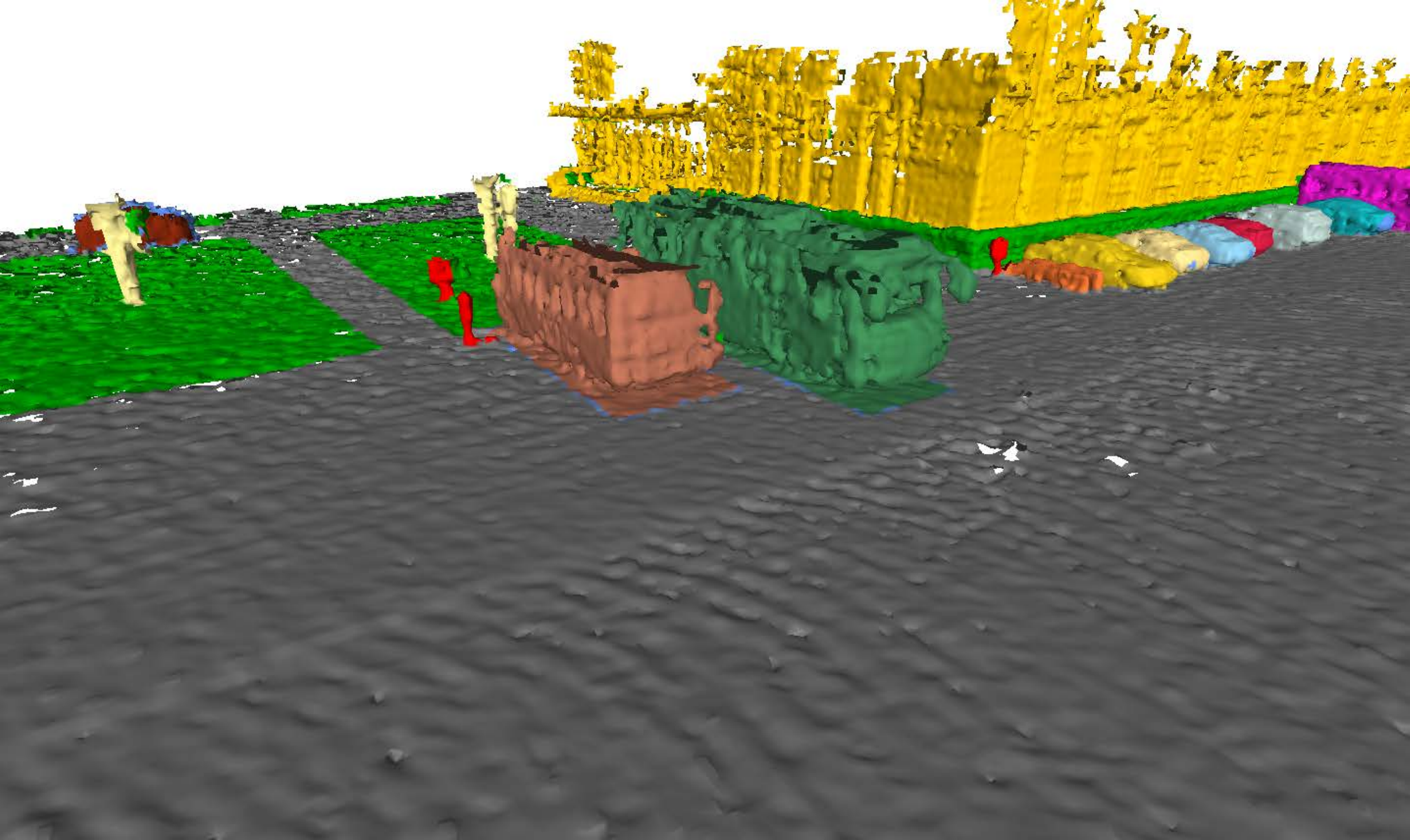}}
\vspace{-0.1em}
\caption{The visualization of our semantic mapping results (The first row and the third row) and panoptic mapping results (The second row and the fourth row) on different sequences of different datasets. "SK-00" means sequence 00 of the SemanticKITTI dataset, "SP-00" means sequence 00 of the SemanticPOSS dataset, so on and so forth.}
\vspace{-0.6cm}
\label{fig:different-data}
\end{figure*}
\section{EXPERIMENTS}
We propose an implicit semantic mapping framework which supports four working modes: incremental semantic mapping, incremental panoptic mapping, batch semantic mapping and batch panoptic mapping. Unless explicitly stated, experiment results reported in this section are from batch mode by default.
\subsection{Experimental Setup}
\textbf{Datasets}. We evaluate our method on two public outdoor LiDAR datasets. One is SemanticKITTI \cite{behley2019iccv}, which provide the labels of semantic segmentation and panoptic segmentation for every point cloud. The other is SemanticPOSS \cite{pan2020semanticposs}, which contains 2988 various and complicated LiDAR scans with large number of dynamic object instances. 

\textbf{Metrics for Mapping Quality}. Our evaluation metric is based on Chamfer Distance, which is a metric commonly used in computer vision community to quantify the dissimilarity between two point sets or shapes. Inspired by \cite{herb2021lightweight}, we utilize Semantic Chamfer Distance (SCD) to evaluate the quality of our semantic mesh. We compute the distance between the points that belong to the same classes. To compute the distance for a pair of reconstructed and ground truth mesh, we randomly sample 5 million points $R$ and $T$ in our reconstructed semantic mesh and ground truth mesh, respectively. Since ground truth mesh is not provided for these two datasets, we define the labeled point cloud as ground truth. For each point $r\in R$ with the class $c_r$ and each point $t \in T$ with the class $c_t$ and , the definition of SCD is:
\begin{equation}
    d_{t\to R}= \mathop{min}_{r\in R, c_t=c_r} \|r-t\|, d_{r\to T}= \mathop{min}_{t\in T, c_r=c_t} \|t-r\| 
\end{equation}

\textbf{Implementation Details.} For the MLPs, all the number of hidden layers of MLPs $p$ is 2. The $s_{cube}$ is 0.1 m and the $\alpha$ is 0.05, the number of sampled points for one LiDAR ray $N$ is 6. Setting $\sigma$ as $1/3$, the lengths of feature embeddings $H_1$ and $H_2$ are 8 and 16, respectively. For SNF, we use feature embeddings of full length to represent semantic information. For PNF, one third of $H_2$ is used to store semantic information and the other is used to store instance information. 

\subsection{Map Quality Evaluation}
\label{sub:mapqu}
\textbf{Qualitative Results}. Fig. \ref{fig:different-scene} shows our semantic reconstruction effects on SemanticKITTI dataset, we compare our method with Suma++\cite{chen2019suma++} and LODE \cite{li2023lode}. Our semantic map is denser compared to Suma++. Although Suma++ removed dynamic objects, it eliminates static vehicles as well, which is not anticipated for certain tasks. However, we successfully preserve static vehicles and only removed dynamic ones. Compared to LODE, our map is more precise and complete in terms of semantics. 
Besides, Fig. \ref{fig:different-data} shows the effectiveness of our semantic and panoptic reconstruction on SemanticKITTI dataset and SemanticPOSS dataset.
\begin{table}[h]
\caption{The F-score value of each semantic categories with a 0.25cm error threshold on SemanticKITTI 00. Only the SemanticKITTI 00 is evaluated in the related papers, same configuration is applied here for fair comparison. And our semantic label can be predicted by a vanilla semantic segmentation method, as simple as RangeNet++ \cite{milioto2019rangenet++}.}
\centering
\scalebox{0.95}{
\begin{tabular}{ c c c c c c c}
\hline
  Methods & Road &Side & Sign & Pole & Barrier & Building  \\
\hline

 Kimera(20cm) \cite{rosinol2020kimera}&90.6&75.7&30.3&16.1&39.9&44.2\\
 Kimera(10cm) \cite{rosinol2020kimera}&91.2&77.7&39.8&51.9&45.0&51.1\\
 Kimera(5cm) \cite{rosinol2020kimera}&91.6&79.2&49.4&\textbf{71.4}&51.1&57.7\\
 Lightweight \cite{herb2021lightweight}&89.5&74.1&53.7&58.7&\textbf{53.6}&50.1\\
\hline
 Ours&\textbf{92.8}&\textbf{81.9}&\textbf{86}&59&51.9&\textbf{68.3}\\
 \hline
\end{tabular}
}
\label{tab: kimera}
\vspace{-0.3cm}
\end{table}
\textbf{Quantitative Results.}
For SemanticKITTI dataset, we use the labeled point cloud as ground-truth. To the best of our knowledge, there is no existing algorithms that takes LiDAR-only input to reconstruct a semantic mesh map, we compare our approach against Kimera, which provides a TSDF-based semantic reconstruction and Lightweight \cite{herb2021lightweight}, which is a semantic reconstruction method. Both methods use image sensor as input. For Kimera, we choose a voxel size of 5 cm, 10 cm, 20 cm to evaluate. By using SCD metric, we further calculate the reconstruction metric F-score \cite{knapitsch2017tanks}. TABLE \ref{tab: kimera} shows our quantitative reconstruction result.
\begin{table}[h]
\caption{ Quantitative results of our experiments for map precision on SemanticKITTI and Semantic POSS. We show the metric on semantic map and panoptic map with a 10cm error threshold in the "batch mode" and "incremental mode". "Com" = completion in cm, "ACC"=accuracy in cm, "Ch-L1" = Champer-L1 in cm, "Com.R" = Completion Ratio, "SK" =  SemantiKITTI, "PO" = SemanticPOSS. "SEB"=semantic map in batch mode, "SEI"= semantic map in incremental map, "PAB"=panoptic map in batch mode, "PAI"=panoptic map in incremental mode.}
\scalebox{0.81}{
\begin{tabular}{ c| c| c c c c c }
\hline
 Method &Data& Com $\downarrow$ & Acc $\downarrow$ &Ch-L1 $\downarrow$&Com.R $\uparrow$ &F-score $\uparrow$  \\
\hline
SHINE\_Mapping \cite{zhong2023shine} & SK& \textbf{4.9}&6.9&5.9&\textbf{92.4}&81.3\\
NeRF\_LOAM \cite{deng2023nerf} & SK& 6.9&5.1&6.0&87.7&87.8\\
Ours (SEB) &SK&7.1&\textbf{4.3}&\textbf{5.7}&90.7&\textbf{91.0}\\
Ours (PAB)  &SK & 5.8&6.9&6.4&91.3&81.1\\
Ours (SEI)&SK&12.9&5.2&9.1&79.3&81.8\\
Ours (PAI)&SK&8.7&6.4&7.6&75.2&76.9\\
\hline
SHINE\_Mapping \cite{zhong2023shine}&PO&7.4&7.2&7.3&85.4&79.1\\
NeRF\_LOAM \cite{deng2023nerf} &PO& 17.1&6.0&11.6&74.3&78.6\\
Ours (SEB) &PO& \textbf{7.2}&7.3&\textbf{7.2}&\textbf{87.1}&79.3\\
Ours (PAB) &PO& 13.7&\textbf{5.3}&9.5&75.2&\textbf{81.4}\\
Ours (SEI)&PO&10.1&7.3&8.7&74.3&73.5\\
Ours (PAI)&PO&13.7&7.6&10.6&70.5&70.4\\
 \hline
\end{tabular}
}
\label{table:copshine}
\vspace{-0.4cm}
\end{table}

Obviously, our result achieves a better result, one of the reason is that LiDAR's range measurement is more precise than a RGB-D sensor. To be fair, we compare our method against SHINE\_Mapping and NeRF-LOAM, which are implicit reconstruction mapping methods using LiDAR data and ground-truth poses as input, in the second experiment. We use commonly used reconstruction metric from SHINE\_Mapping, accuracy, completion, Chamfer-L1 distance, completion ratio and F-score, to evaluate our mapping quality on SemanticKITTI and SemanticPOSS. Since the two datasets dare unable to provide ground-truth mesh, the point cloud is deemed as ground-truth here. As shown in TABLE \ref{table:copshine}. Overall, we can see that our method achieves better average performance compared to the other two methods. However, Shine\_Mapping narrowly outperforms our method in some metrics related to map completion in the SemanticKITTI dataset. This is because our method is able to eliminate dynamic objects, while they are still existing inside ground truth, it has an impact on the map completion score. In addition, the map is continuously updated based on new input data in the incremental mapping mode, which introduces some level of instability or inconsistency due to the dynamic nature of the updates. In contrast, batch mapping mode reconstructs the map in a single step based on the entire dataset, leading to a more consistent and stable map representation. Consequently, while incremental mapping may exhibit slightly lower map quality compared to batch processing, it is better suited for real-time applications where the map needs to be updated rapidly to accommodate environmental changes. And as discussed in \ref{sub:mapqu}, our mapping result includes semantic information, which also mark our improvement compared to SHINE\_Mapping and NeRF-LOAM.  
\begin{figure}[t!]
    \centering
    \vspace{10pt}

\captionsetup{position=b}

\captionsetup[subfigure]{labelformat=empty}
\subfloat[Submap1\label{Psub21:gt}]{\includegraphics[width=.22\linewidth]{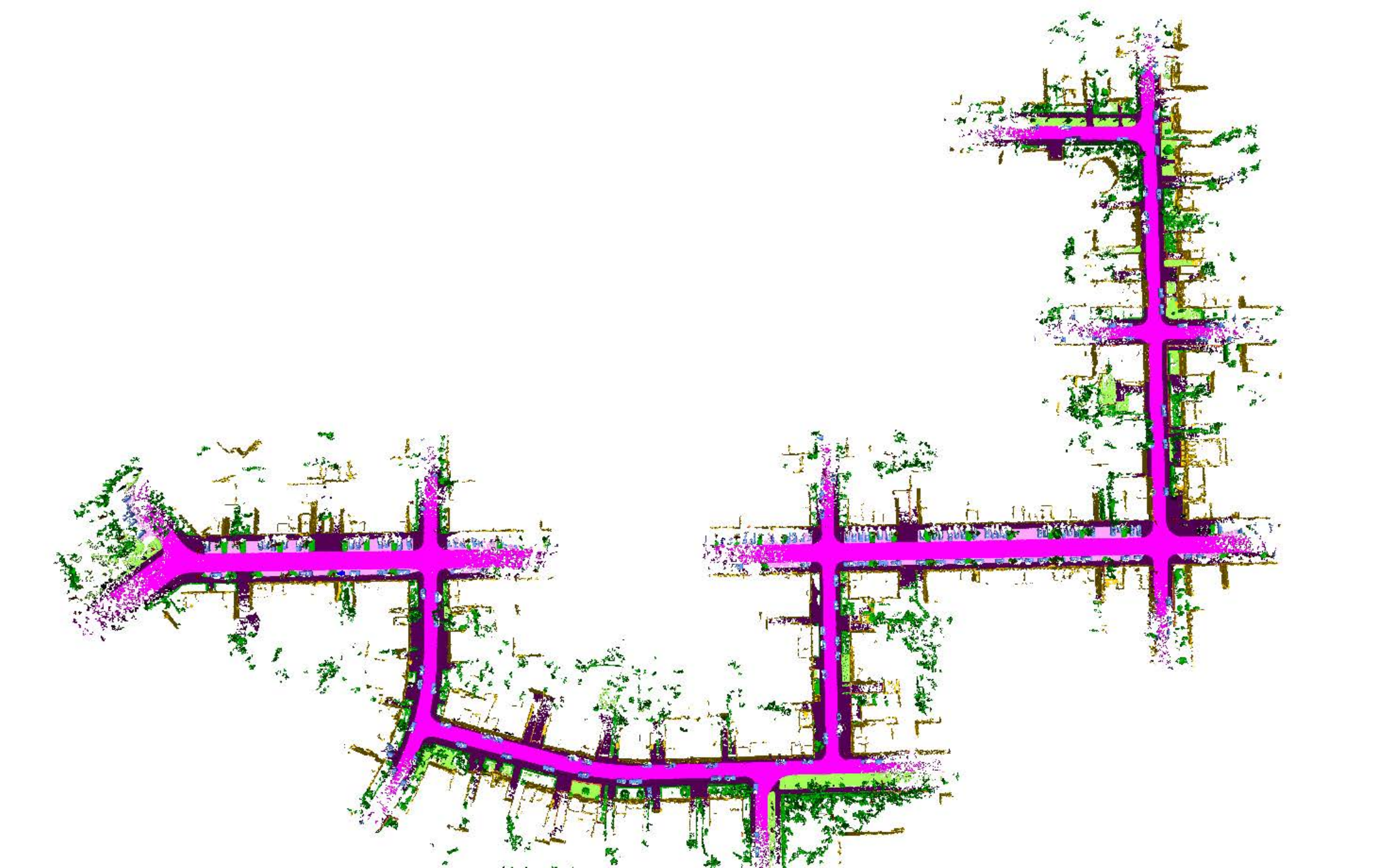}} 
\subfloat[Submap2]{\includegraphics[width=.22\linewidth]{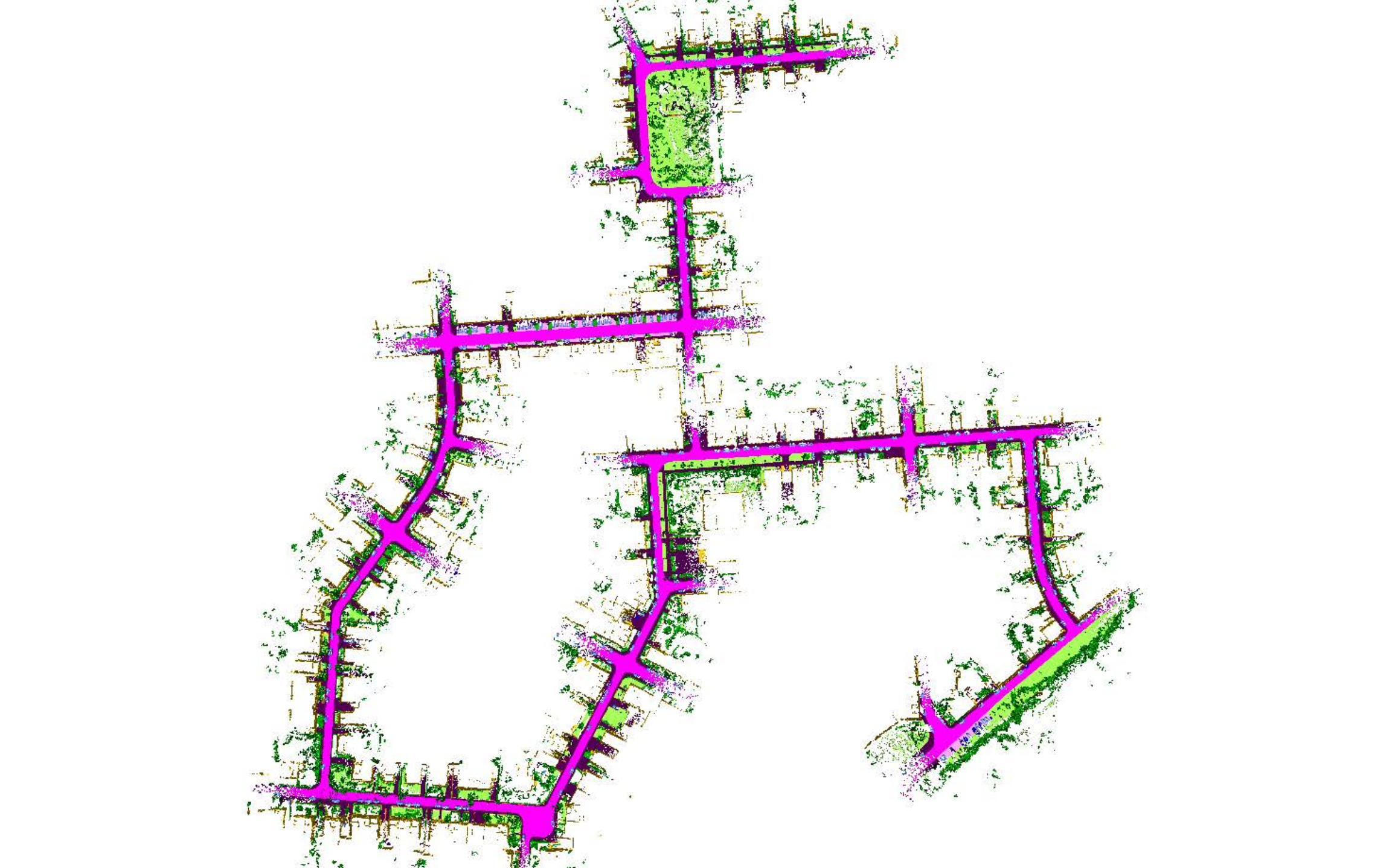}}    
\subfloat[Submap3]{\includegraphics[width=.22\linewidth]{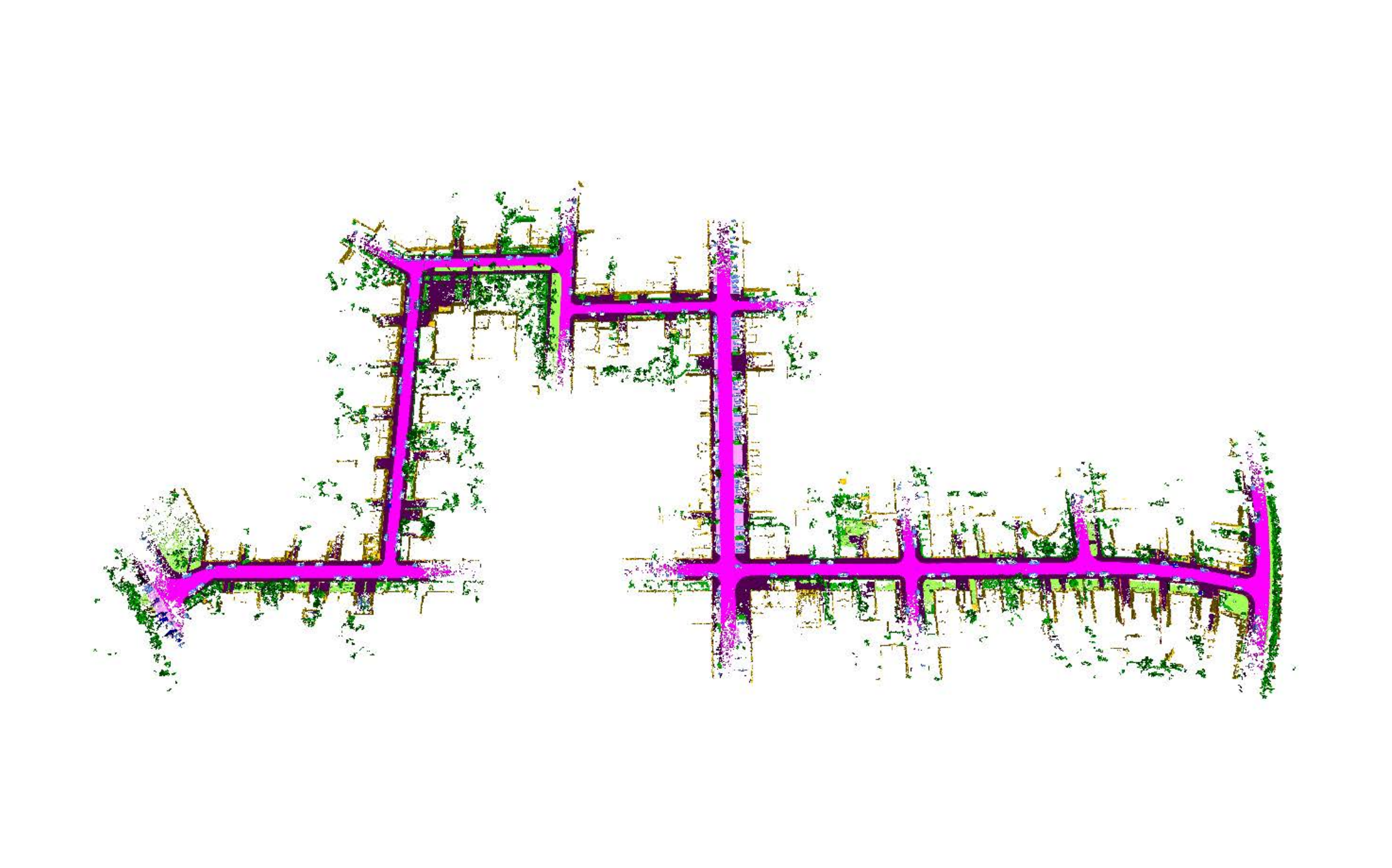}} 
\subfloat[Submap4]{\includegraphics[width=.22\linewidth]{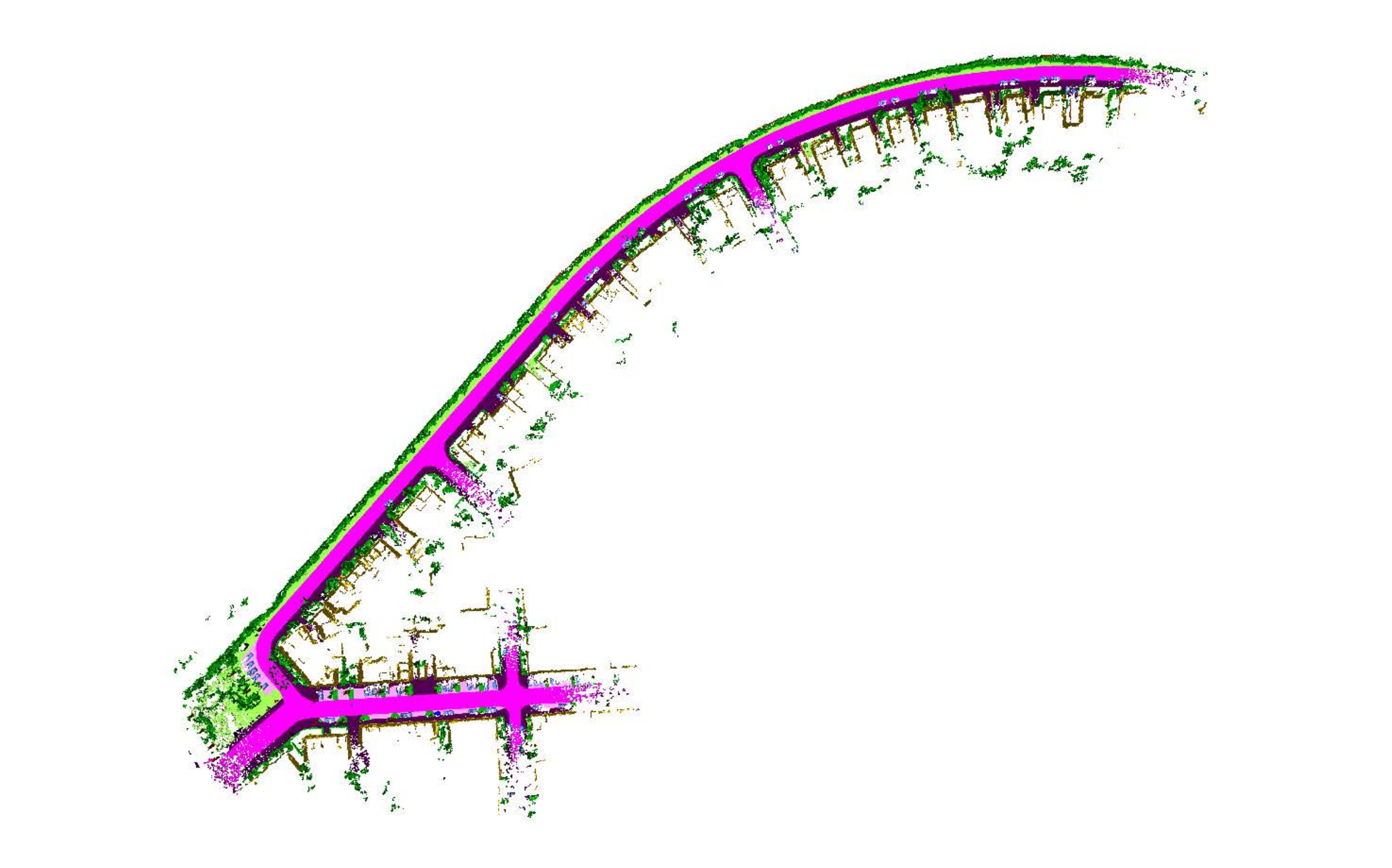}}.  
\subfloat[The whole map]{\includegraphics[width=1\linewidth]{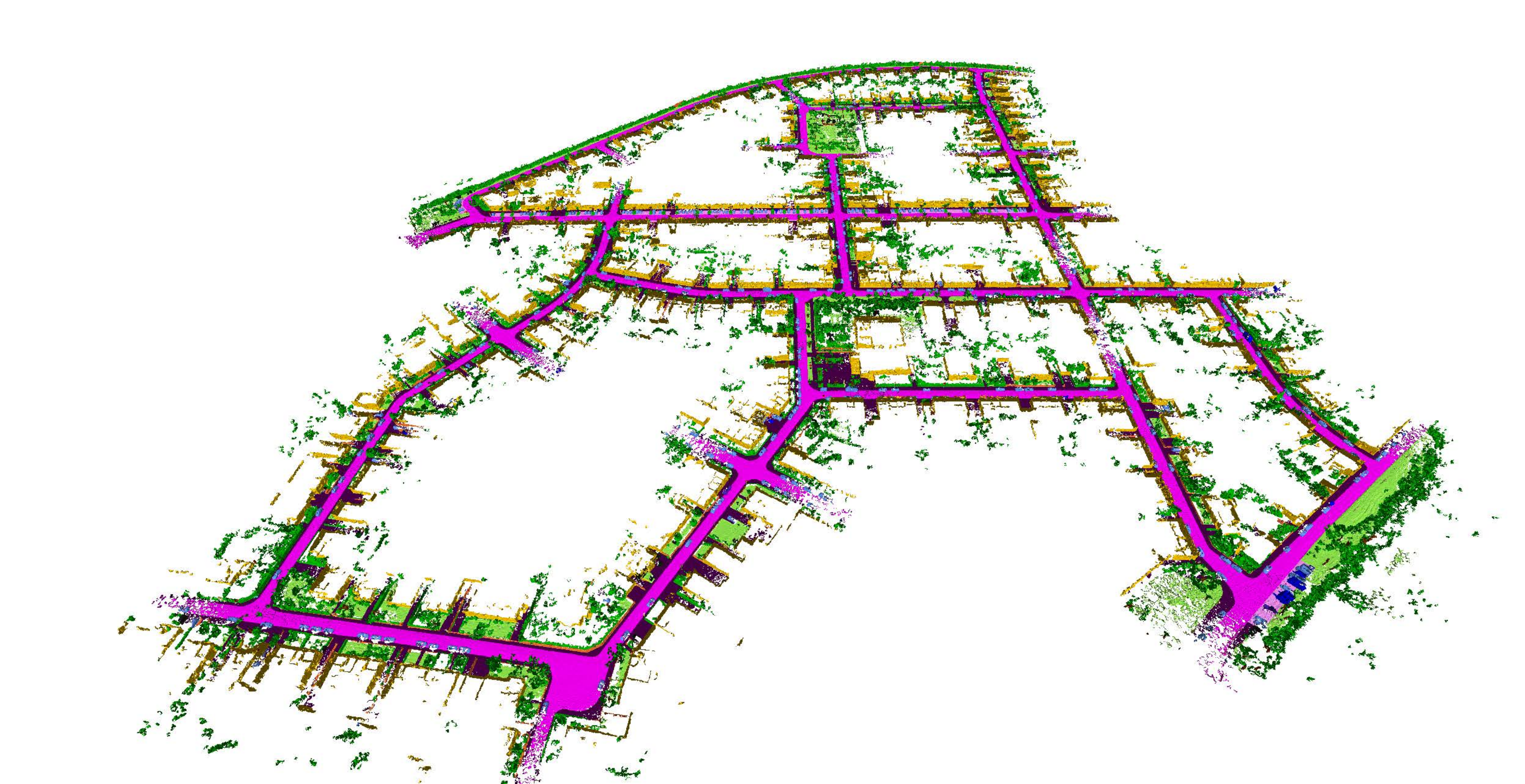}} 

    \caption{The visualization of sub-maps and the whole map}
    \label{fig:kitti}
    \vspace{-0.4cm}
\end{figure}
\subsection{Results and Analysis of Map Merge}
We also provide a demo experiment for map merge, which is conducted on SemanticKITTI sequence 00, and the qualitative result is shown on Fig. \ref{fig:kitti}. We divide the whole environment SemanticKITTI sequence 00 into four sections to build. From the visualization result in Fig. \ref{fig:kitti}, semantic submap 1 to semantic submap 4 are successfully built, and the final map is generated via aligning these four submaps. This validates our framework based on the neural semantic fields is promising to extend to urban-level mapping scenarios. The number of overlapping frames between adjacent submaps is examined with 1, 5, and 10. Increasing the amount of overlapping frames gradually improve the fusion results.


\subsection{Evaluation for Dynamic Scenario}
\begin{figure}[h!]
    \centering
    \subfloat{\includegraphics[width=.48\linewidth]{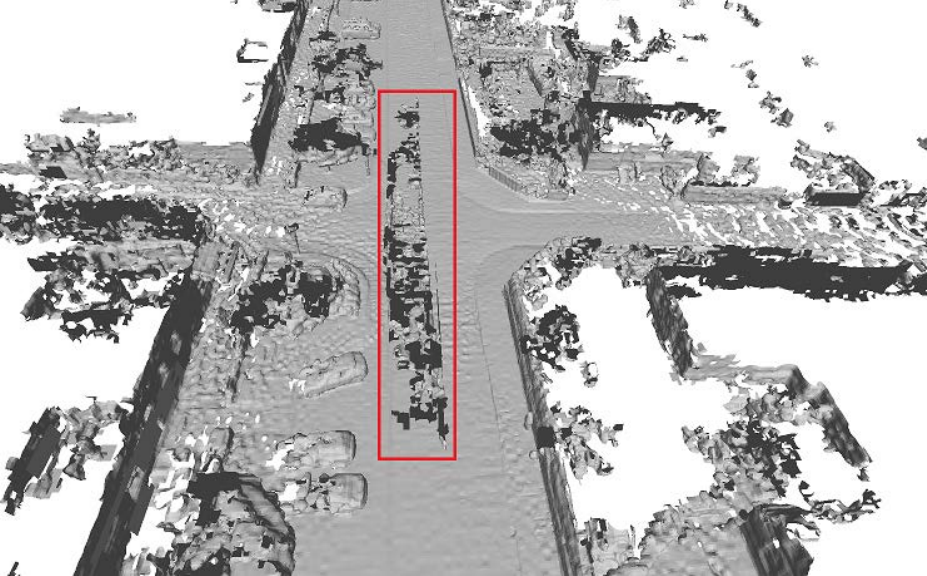}}
     \subfloat{\includegraphics[width=.48\linewidth]{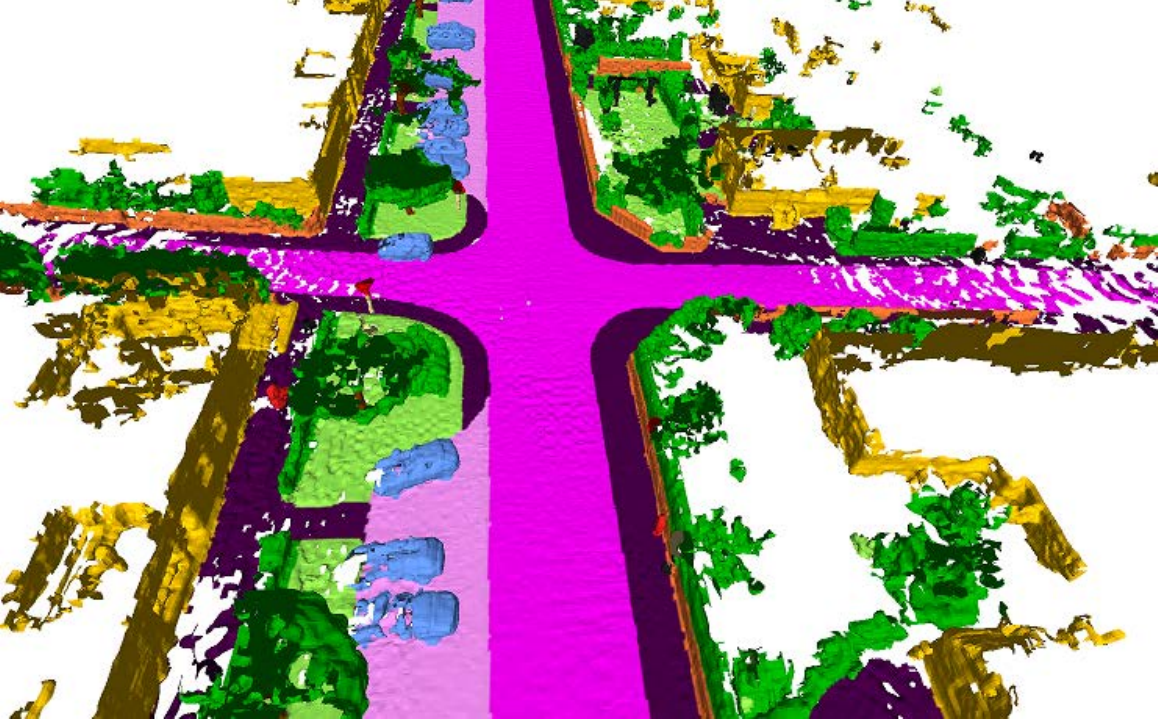}}
    \caption{\textbf{Illustration of the superiority of our LISNeRF for dynamic scenario from SemanticKITTI.} Left image is generated from SHINE\_Mapping, whose map is polluted by moving vehicles (ghost trails shown within the red box). By enrolling semantic information, our map (right image) successfully eliminates dynamic objects and obtains smooth road surface.}
    \label{fig:dy}
    \vspace{-0.5cm}
\end{figure}

In highly dynamic environments, it is challenging to generate a static map without the interference of dynamic elements. As shown in Fig. \ref{fig:dy}, our method is able to eliminate certain types of the dynamic objects by introducing semantic label into the mapping procedure. In practice, we distinguish the dynamic objects from static backgrounds using labels from RangeNet++ \cite{milioto2019rangenet++}. Triangle mesh corresponding to the dynamic objects is filtered correspondingly during the mapping procedures.

\subsection{Comparison between Two Label Strategies }
\begin{figure}[h!]

\captionsetup[subfigure]{labelformat=empty}

    \centering
    \subfloat{\includegraphics[width=.48\linewidth]{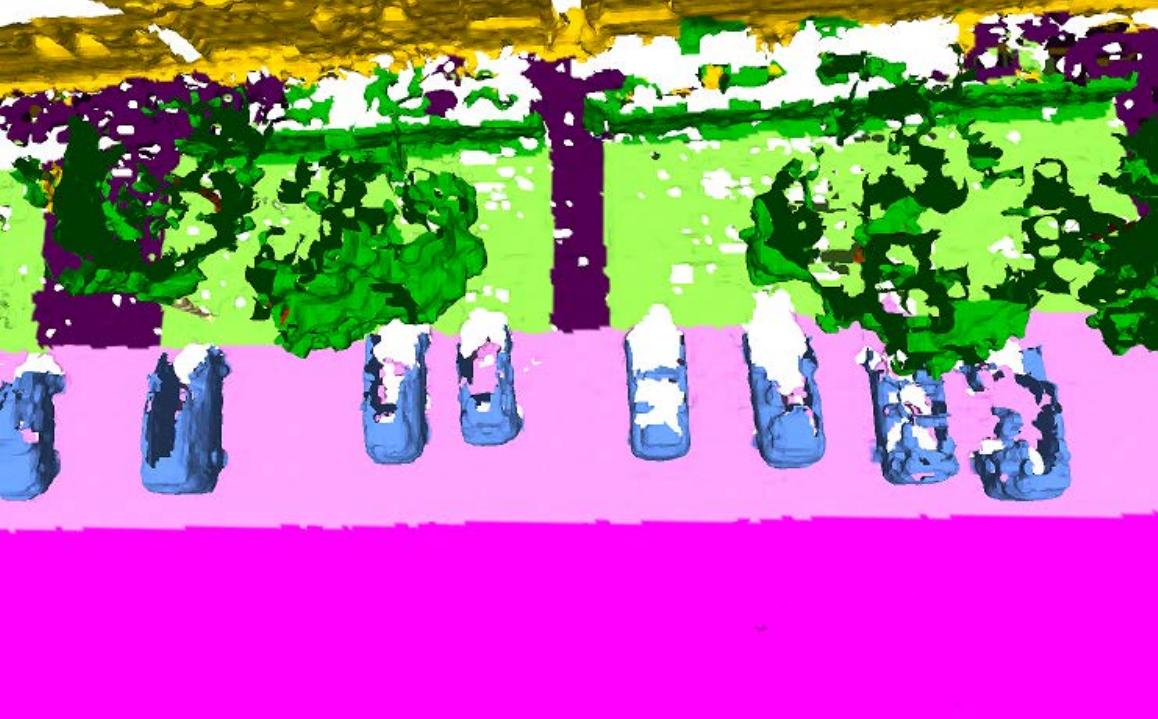}}
    \hspace{0.1em}
    \subfloat{\includegraphics[width=.48\linewidth]{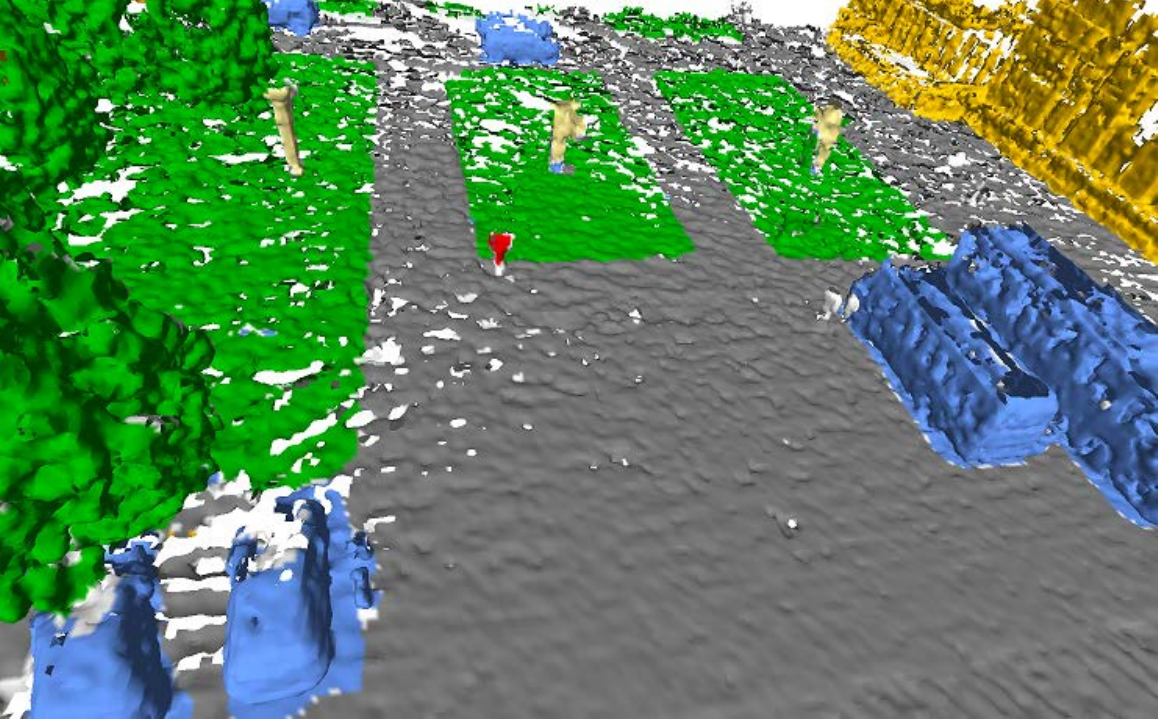}}\\ 
    \vspace{-0.5em} 
    \subfloat[SemantiKITTI-05]{\includegraphics[width=.48\linewidth]{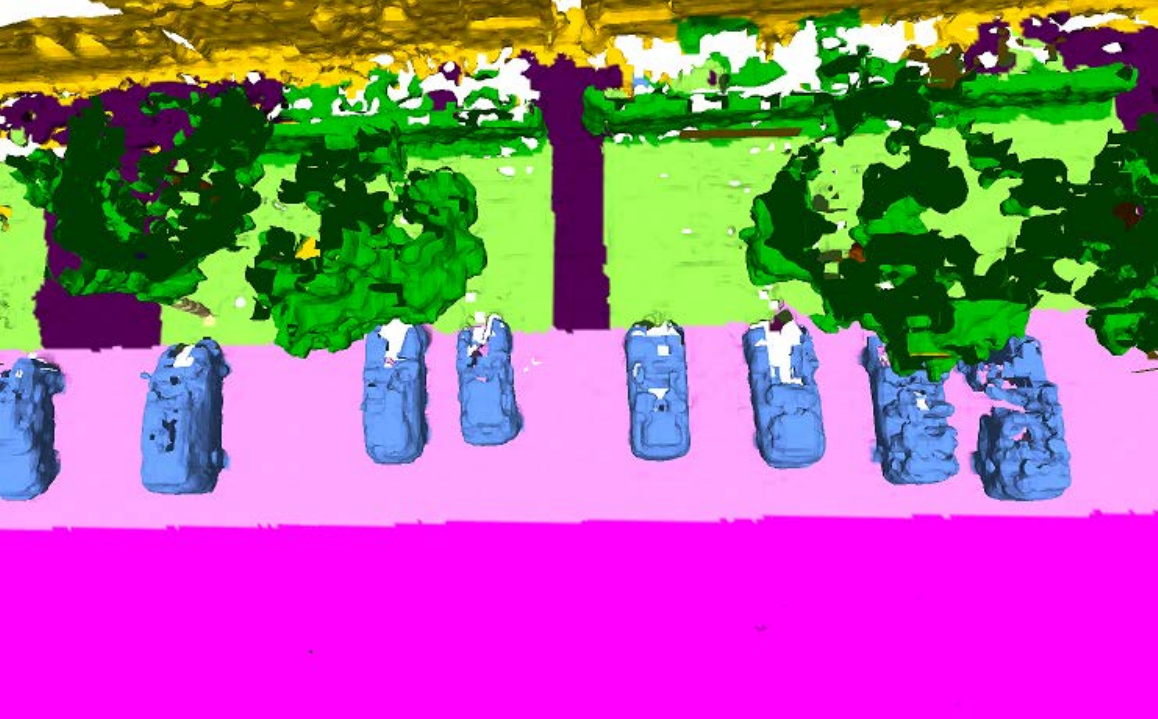}}
    \hspace{0.1em}
    \subfloat[SemanticPOSS-01]{\includegraphics[width=.48\linewidth]{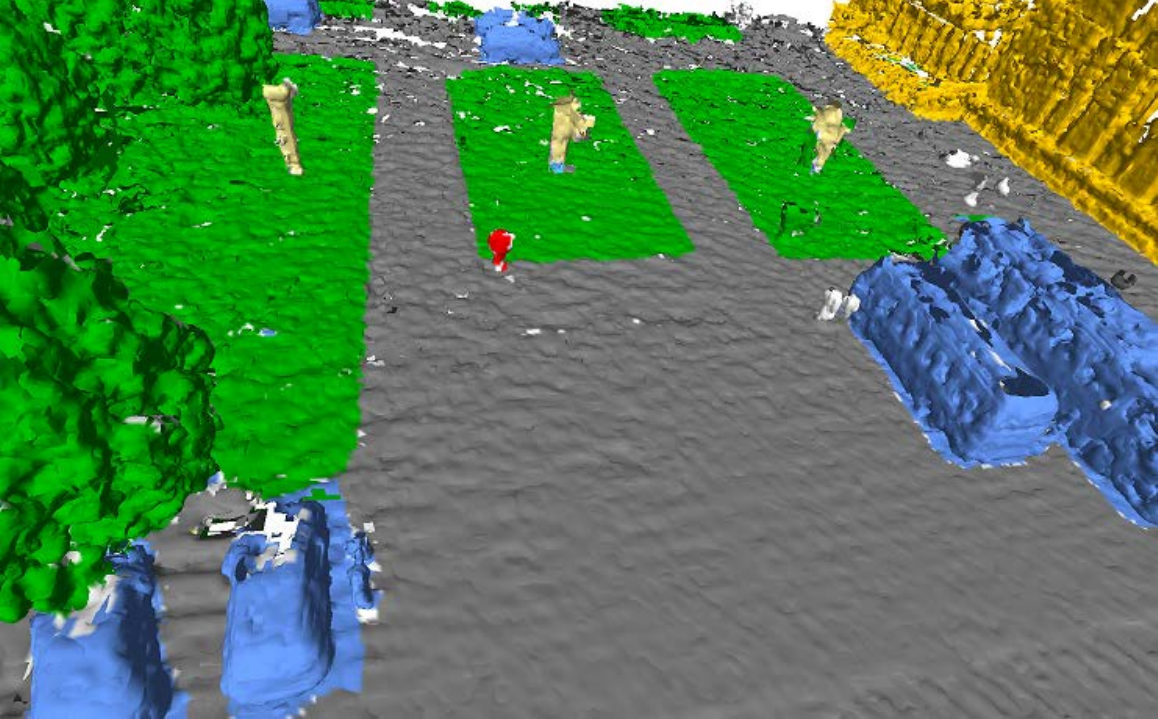}}
    \caption{\textbf{Test of sampling strategies on SemanticKITTI and SemanticPOSS dataset.} First row shows mapping result when semantic labels are assigned to all sampled point and second row is mapping result generated from surface-only labeling strategy. We can see that the semantic maps of the second row is significantly denser than those of the first row. 
    }
\label{fig:dif-label}
\vspace{-0.3cm}
\end{figure}

As mentioned in \ref{item:snf}, we only assign the same semantic labels to the sampled points near the objects surface in the octree building stage. To demonstrate the effectiveness of our strategy, the result of our semantic mapping is shown in Fig. \ref{fig:dif-label}. When we assign semantic labels to all sampled points along the LiDAR beam, the free space is also assigned with a semantic label, which misleads the training of MLPs. This strategy generates more holes and errors for objects mesh in the inferring stage. This phenomenon is largely improved by only train MLPs to remember the semantic status of objects' surface area. 


\section{CONCLUSION}
This paper proposes a novel method for large-scale implicit mapping with semantic information. We implicitly store geometry  and semantic information in vertices of octree nodes using hashing tables, then we leverage multiple MLPs to decode the feature embeddings to SDF value and semantic categories. We evaluate our method on two datasets and the results show that our method achieves decent mapping quality and efficiency. Besides, our framework is able to realize incremental mode and batch-based mode for semantic mapping and panoptic mapping. 
For large-scale mapping, we also validate a map merge strategy and dynamic objects elimination method, which are helpful for multi-agents collaborative mapping in the real-world tasks. In the future, we plan to incorporate odometry and loop closure techniques to LISNeRF to develop a full-stack SLAM system. 

\bibliographystyle{IEEEtran.bst}
\bibliography{bib.bib}

\end{document}